\documentclass{article}

\PassOptionsToPackage{numbers, compress}{natbib}
 \usepackage[final]{nips_2016}

\usepackage[utf8]{inputenc} % allow utf-8 input
\usepackage[T1]{fontenc}    % use 8-bit T1 fonts
\usepackage{url}            % simple URL typesetting
\usepackage{booktabs}       % professional-quality tables
\usepackage{amsfonts}       % blackboard math symbols
\usepackage{nicefrac}       % compact symbols for 1/2, etc.
\usepackage{microtype}      % microtypography
\usepackage{amssymb}
\usepackage{epsfig}
\usepackage{amsfonts}
\usepackage{mathrsfs}
\usepackage{amsmath}
\usepackage{graphicx}
\usepackage{color}
\usepackage{caption}
\usepackage{picinpar}
\usepackage{booktabs}
\usepackage{float}
\usepackage{url}
\usepackage{dsfont}
\usepackage{multirow}
\usepackage{subfloat}
\usepackage{slashbox}
\usepackage{array}
\usepackage{subfigure}
\usepackage[none]{hyphenat}
\title{Robust Stochastic Configuration Networks with Kernel Density Estimation }

\author{
  Dianhui~Wang\thanks{Corresponding author. }, ~~Ming Li \\ \\
  Department of Computer Science and Information Technology\\
  La Trobe University,
  Melbourne, VIC 3086, Australia \\
  \texttt{Email:dh.wang@latrobe.edu.au} \\
}

\begin{document}

\maketitle

\begin{abstract}
Neural networks have been widely used as predictive models to fit data distribution, and they could be implemented through learning a collection of  samples. In many applications, however, the  given  dataset may contain  noisy samples or outliers which may result in  a poor learner model in terms of generalization.  This paper contributes to a development of robust stochastic configuration networks (RSCNs) for resolving uncertain data regression problems. RSCNs are built on original stochastic configuration networks with weighted least squares method for evaluating the output weights, and the input weights and biases are incrementally and randomly generated by satisfying with a set of inequality constrains. The kernel density estimation (KDE) method is employed to set the penalty weights for each training samples, so that some negative impacts, caused by noisy data or outliers, on the resulting learner model can be reduced.  The alternating optimization technique is applied for updating a RSCN model with improved penalty weights computed from the kernel density estimation function. Performance evaluation is carried out by  a function approximation, four benchmark datasets and a case study on engineering application. Comparisons to other robust randomised neural modelling techniques, including the probabilistic robust learning algorithm for neural networks with random weights and improved RVFL networks, indicate that the proposed RSCNs with KDE perform favourably and demonstrate good potential for real-world applications.

%\textbf{Keywords:} Stochastic configuration networks, robust data regression, randomized algorithms, kernel density estimation, alternating optimization techniques
\end{abstract}

\section{Introduction}
For many real-world applications, sample data collected from various sensors may be contaminated by some noises or outliers \cite{Hampel2011robust}, which makes troubles for building neural networks with sound generalization. Over the past years, robust data modelling techniques have received considerable attention in the field of applied statistics \cite{Hampel2011robust,Huber1981robust,Leroy1987robust} and machine learning \cite{Chen1994,Chuang2002,El2013random,Liano1996,Suykens2002}. It is well known that the cost function plays an important role in robust data modelling. In \cite{Chen1994}, the M-estimator and Hampels hyperbolic tangent estimates were employed in the cost function, aiming to alleviate the negative impacts of outliers on the modelling performance. Under an assumption that the additive noise of the output follows Cauchy distribution, the mean log squared error was used as a cost function in \cite{Liano1996}. In \cite{El2013random}, a robust learning algorithm based on the M-estimator cost function with random sample consensus was proposed to deal with outliers, and this algorithm has been successfully applied in computer vision and image processing \cite{Meer1991robust,Torr2000mlesac,Zhuang1992highly}. Besides these methods mentioned above, some results on robust data regression using support vector machine (SVM) have been reported in \cite{Chuang2002,Suykens2002}, where SVM-based approaches demonstrate some limits to handle uncertain data regression problems with higher level outliers.

Back-propagation algorithms for training neural networks suffer from many shortcomings, such as learning parameter setting, slow convergence and local minima. Thus, it is useful to develop advanced learning techniques for resolving data regression problems, in particular, for stream data or online data modelling tasks. With such a background, randomized methods for training neural networks have been developed in the last decades \cite{Igelnik1995, Broomhead1988,Pao1992}. Readers may refer to a recently published survey paper  for more details about some milestones on this topic \cite{ScardapaneandWang2017}. Studies on the robust data modelling techniques based on Random Vector Functional-link (RVFL) networks have been reported in \cite{Cao2015,Dai2015}. Specifically, a hybrid regularization model with assumption on the sparsity of outliers was used in training process, and a probabilistic robust learning algorithm for neural networks with random weights (PRNNRW) was proposed in \cite{Cao2015}. However, some learning parameters used in PRNNRW must be set properly and this is quite difficult to be done in practice. In \cite{Dai2015}, an improved version of RVFL networks built by using a KDE-based weighted cost function was suggested.
Unfortunately, the significance of the scope setting of the random weights and biases for RVFL networks has not been addressed. In \cite{LiandWang2016}, we looked into some practical issues and common pitfalls of RVFL networks, and clearly revealed  the impact of the scope setting on the modelling performance of  RVFL networks. Our findings reported in \cite{LiandWang2016} motivates us to further investigate the robust data regression problem using an advanced randomized learner model, termed as Stochastic Configuration Networks (SCNs), which are built incrementally by assigning the random weights and biases with a supervisory mechanism \cite{WangandLi_SCN}.

This paper aims to develop a robust version of SCNs for uncertain data regression. Based on the construction process of SCNs, we utilise  a weighted least squares objective function for evaluating the output weights of SCNs, and the resulting approximation errors from the present SCN model are used to incrementally configure the hidden nodes with constrained random parameters. During the course of building RSCNs, the penalty weights representing the degree of contribution of individual data samples to the objective function are  updated according to a newly constructed KDE function. In this work, an alternating optimization (AO) technique is employed to implement  the RSCN model. Our proposed algorithm, termed as RSC-KDE, is evaluated by using a function approximation, four benchmark datasets with different levels of artificial outliers, and an engineering application \cite{Dai2015}. Experimental results  indicate that the proposed RSC-KDE outperforms other existing methods in terms of effectiveness and robustness.

The remainder of paper is organized as follows: Section 2 briefly reviews stochastic configuration networks and the  kernel density estimation  method.  Section 3 details our proposed RSC-KDE algorithm. Section 4 reports experimental results with comparisons and discussion. Section 5 concludes this paper with some remarks.
\section{Revisit of Stochastic Configuration Networks}
This section reviews our proposed SCN framework, in which the input weights and biases are randomly assigned in the light of a supervisory mechanism, and the output weights are evaluated by solving a linear least squares problem. More details about SCNs can be read in \cite{WangandLi_SCN}.

Let $\Gamma:=\{g_1, g_2, g_3...\}$ be a set of real-valued functions, span$(\Gamma)$ denote a function space spanned by $\Gamma$; $L_{2}(D)$ denote the space of all Lebesgue measurable functions $f=[f_1,f_2,\ldots,f_m]:\mathds{R}^{d}\rightarrow \mathds{R}^{m}$ defined on $D\subset \mathds{R}^{d}$, with the $L_2$ norm defined as
\begin{equation}\label{multiple_lp}
  \|f\|:=\left(\sum_{q=1}^{m}\int_{D}|f_q(x)|^2dx\right)^{1/2}<\infty.
\end{equation}
The inner product of $\phi=[\phi_1,\phi_2,\ldots,\phi_m]:\mathds{R}^{d}\rightarrow \mathds{R}^{m}$ and $f$ is defined as
\begin{equation}\label{multiple_inner}
  \langle f,\phi\rangle:=\sum_{q=1}^{m}\langle f_q,\phi_q\rangle=\sum_{q=1}^{m}\int_{D}f_q(x)\phi_q(x)dx.
\end{equation}
In the special case that $m=1$, for a real-valued function $\psi:\mathds{R}^{d}\rightarrow \mathds{R}$ defined on $D\subset \mathds{R}^{d}$, its $L_2$ norm becomes $ \|\psi\|:=(\int_{D}|\psi(x)|^2dx)^{1/2}$, while the inner product of $\psi_1$ and $\psi_2$ becomes $\langle \psi_1,\psi_2\rangle=\int_{D}\psi_1(x)\psi_2(x)dx$.

Given a target function $f:\mathds{R}^{d}\rightarrow \mathds{R}^{m}$, suppose that we have already built a single layer feed-forward network (SLFN) with $L-1$ hidden nodes, i.e, $f_{L-1}(x)=\sum_{j=1}^{L-1}\beta_jg_j(w_j^\mathrm{T}x+b_j)$ ($L=1,2,\ldots$, $f_0=0$), $\beta_j=[\beta_{j,1},\ldots,\beta_{j,m}]^\mathrm{T}$, and the current residual error, denoted as $e_{L-1}=f-f_{L-1}=[e_{L-1,1},\ldots,e_{L-1,m}]$, does not reach an acceptable tolerance level. The framework of SCNs provides an effective solution for how to add $\beta_L$, $g_L$ ($w_L$ and $b_L$) leading to $f_{L}=f_{L-1}+\beta_Lg_L$ until the residual error $e_L=f-f_L$ falls into an expected tolerance $\epsilon$. The following Theorem 1 restates the universal approximation property of SCNs (corresponding to Theorem 7 in \cite{WangandLi_SCN}).\vspace{1.8mm}

\textbf{Theorem 1.} Suppose that span($\Gamma$) is dense in $L_2$ space and $\forall g\in \Gamma$, $0<\|g\|<b_g$ for some $b_g\in \mathds{R}^{+}$. Given $0<r<1$ and a nonnegative real number sequence $\{\mu_L\}$ with $\lim_{L\rightarrow+\infty}\mu_L=0$ and $\mu_L\leq (1-r)$.  For $L=1,2,\ldots$, denoted by
\begin{equation}
\delta_{L}=\sum_{q=1}^{m}\delta_{L,q}, \delta_{L,q}=(1-r-\mu_L)\|e_{L-1,q}\|^2, q=1,2,\ldots,m.
\end{equation}
If the random basis function $g_L$ is generated with the following constraints:
\begin{equation}\label{step3}
\langle e_{L-1,q},g_L\rangle^2\geq b_g^2\delta_{L,q}, q=1,2,\ldots,m,
\end{equation}
and the output weights are evaluated by
\begin{equation}\label{step4}
[\beta^{*}_1, \beta^{*}_2,\ldots,\beta^{*}_{L}]=\arg \min_{\beta}\|f-\sum_{j=1}^{L}\beta_jg_j\|.
\end{equation}
Then, $\lim_{L\rightarrow +\infty}\|f-f_L\|=0,$ where $f_L=\sum_{j=1}^{L}\beta^{*}_{j}g_j$, $\beta^{*}_{j}=[\beta^{*}_{j,1},\ldots,\beta^{*}_{j,m}]^{\mathrm{T}}$.\vspace{1.8mm}

The construction process of SCNs starts with a small sized network, then incrementally adds hidden nodes followed by computing the output weights. This procedure keeps going on until the model meets some certain termination criterion. Some remarkable merits of our SCNs can be summarized as follows : (i) There is no requirement on any prior knowledge about the architecture of the constructed network for a given task; (ii) The scope of the random weights and biases is adjustable and automatically determined by the data rather than a fixed setting from end-users; and (iii) The input weights and biases are randomly assigned with an inequality constraint, which can guarantee the universal approximation property.

\textbf{Remark 1:} In the past two decades, randomized methods for training neural networks suffer from some misunderstandings on the constraint of random assignment (i.e., randomly assign the input weights and biases in a fixed interval, even any intervals, or no specification at all). Unfortunately, in many published works the authors blindly and wrongly use the randomness in constructing randomized learner models and mindlessly made some misleading statements without any scientific justification \cite{LiandWang2016}. It should be pointed out that the universal approximation theorems for RVFL networks established in \cite{Igelnik1995} are fundamental and significant to the randomized learning theory. However, these theoretical results cannot provide us with practical and useful guidance on structure and learning parameter settings, and algorithm design and implementation aspects. Our proposed SCN framework firstly touches the base of randomized learning techniques, and draws researches on this topic into right tracks through proper uses of constrained random parameters. Indeed, the inequalities (\ref{step3}) proposed originally in \cite{WangandLi_SCN} are essential to ensure the universal approximation property. From algorithm implementation perspectives, a scheme to prevent SCNs from over-fitting must in place. As usually done in machine learning, the performance over a validation set can be used to terminate the learning process.

At the end of this section, we briefly introduce a kernel density estimation method for weighting contributions of each training data in the learning process. Basically, a kernel density estimator computes a smooth density estimation from data samples by placing on each sample point a function representing its contribution to the density. Readers can refer to \cite{kde1} and \cite{kde2} for more details on the kernel density estimation (KDE) method.

Based on KDE method, the underlying probability density function of a random variable $\eta$ can be estimated by
\begin{equation}\label{kde_def}
  \Phi(\eta)=\sum_{k=1}^{N}\rho_kK(\eta-\eta_k),
\end{equation}
where $K$ represents a kernel function (typically a Gaussian function) centered at the data points $\eta_k$, and $\rho_k$ are weighting coefficients (uniform weights are commonly used, i.e., $\rho_k=1/N$, $k=1,2,\ldots,N$).

\section{Robust Stochastic Configuration Networks}
Robust data regression seeks for a capable learner model that can successfully learn a true distribution from uncertain data samples. This is very important for industrial applications, where the collected data samples are always contaminated by outliers caused by the failure of measuring or transmission devices or unusual disturbances. This section details the development of robust stochastic configuration networks (RSCNs). For a target function $f:\mathds{R}^{d}\rightarrow \mathds{R}^{m}$, given a training dataset with inputs $X=\{x_1,x_2,\ldots,x_N\}$, $x_i=[x_{i,1},\ldots,x_{i,d}]^\mathrm{T}\in \mathds{R}^{d}$ and outputs $T=\{t_1,t_2,\ldots,t_N\}$, where $t_i=[t_{i,1},\ldots,t_{i,m}]^\mathrm{T}\in \mathds{R}^{m}$, $i=1,\ldots,N$, a RSCN model can be built by solving a weighted least squares (WLS) problem, that is,
\begin{equation}\label{wls}
  \min_{\beta,\theta}\sum_{i=1}^{N}\theta_i\|\sum_{j=1}^L\beta_jg(w_j,b_j,x_i)-t_i\|^2,
\end{equation}
where $\theta_i\geq 0$ ($i=1,2,\ldots,N$) is the $i$-th penalty weight, representing the contribution of the corresponding sample to the objective function (\ref{wls}). $G_L(x)=\sum_{j=1}^L\beta_jg(w_j,b_j,x)$ is a SCN, in which $g$ is the activation function and $L$ is the number of hidden nodes, $w_j$, $b_{j}$ are the input weights and biases that are stochastically configured according to Theorem 1, and $\beta_j$ represents the output weights.

Generally speaking, the penalty weights $\theta_i$ ($i=1,2,\ldots,N$) can be determined according to the reliability of the sample $x_i$. It is easy to understand that a higher reliability means more trust in the data that correctly represents the process behavior, and a lower reliability indicates less confidence on the sample that may be an outlier or noisy one. Thus, decreasing (increasing) the penalty weights of training samples with lower (higher) reliability can eliminate or even remove negative impacts on the learner model building.

A logical thinking to combine the original SCN framework with the WLS-based learning is to use a weighted version of the model's residual error in the process of building SCNs. In other words, a RSCN model can be incrementally built by stochastically configuring the hidden parameters based on a redefined constraint (\ref{step3}), and evaluating the output weights by using the WLS solution of (\ref{wls}). Given training samples $X=\{x_1,x_2,\ldots,x_N\}$, $x_i=[x_{i,1},\ldots,x_{i,d}]^\mathrm{T}\in \mathds{R}^{d}$, denoted by $e_{L-1}(X)=[e_{L-1,1}(X),e_{L-1,2}(X),\ldots,e_{L-1,m}(X)]^\mathrm{T}\in \mathds{R}^{N\times m}$, where $e_{L-1,q}(X)=[e_{L-1,q}(x_1),\ldots,e_{L-1,q}(x_N)]\in \mathds{R}^N$, $q=1,2,\ldots,m$. Let $h_L(X)=[g_{L}(w_L^\mathrm{T}x_1+b_L),\ldots,g_{L}(w_L^\mathrm{T}x_N+b_L)]^\mathrm{T}$
be the output vector of the new hidden node for each input $x_i$, $i=1,2,\ldots,N$. Then, we can obtain the current hidden layer output matrix, $H_L=[h_1,h_2,\ldots,h_L]$.

According to (\ref{wls}), a weighted form of $e_{L-1}(X)$ can be defined if the penalty weights are available during the constructive process of SCNs. Denoted the weighted $e_{L-1}(X)$ and $h_L(X)$ by $\tilde{e}_{L-1}(X)=[\tilde{e}_{L-1,1}(X),\tilde{e}_{L-1,2}(X),\ldots,\tilde{e}_{L-1,m}(X)]^\mathrm{T}=\Theta e_{L-1}(X)$, and $\tilde{h}_L(X)=\Theta h_L(X)$, respectively, where $\Theta=\mbox{diag}\{\sqrt{\theta_1},\sqrt{\theta_2},\ldots,\sqrt{\theta_N}\}$. Let $\tilde{\xi}_L=\sum_{q=1}^{m}\tilde{\xi}_{L,q}$ and $\tilde{\xi}_{L,q}$ be defined as
\begin{eqnarray}\label{factor1}
\tilde{\xi}_{L,q}=\frac{\Big(\tilde{e}^{\mathrm{T}}_{L-1,q}(X)\cdot \tilde{h}_L(X)\Big)^2}{\tilde{h}^{\mathrm{T}}_L(X)\cdot \tilde{h}_L(X)}-(1-r-\mu_L)\tilde{e}^{\mathrm{T}}_{L-1,q}(X)\tilde{e}_{L-1,q}(X).
\end{eqnarray}
Based on Theorem 1, the hidden parameters ($w$ and $b$) can be stochastically configured by choosing a maximum $\tilde{\xi}_L$ among multiple tests, subjected to $\tilde{\xi}_{L,q}\geq0$, $q=1,2,\ldots,m$.

Now, the remaining question is how to assign penalty weights $\theta_1,\theta_2,\ldots,\theta_N$ along with the process of building SCNs. Recall that if the probability density function of the residuals can be obtained or estimated, the reliabilities of the samples will be determined. Inspired by the work in [25], we construct a probability density function of the residual error $e_L$ as follows (here $e_L$ is regarded as a random variable)
\begin{equation}
  \Phi(e_L)=\frac{1}{\tau N}\sum_{k=1}^{N}K\left(\frac{\|e_L-e_L(x_k)\|}{\tau}\right),
\end{equation}
where $e_L(x_k)=[e_{L-1,1}(x_k),\ldots,e_{L-1,m}(x_k)]^\mathrm{T}\in \mathds{R}^{m}$, $\tau=1.06\hat{\sigma}N^{-1/5}$ is an estimation window width, $\hat{\sigma}$ is the standard deviation of the residual errors, $K$ is a Gaussian function defined by
\begin{equation}
  K(t)=\frac{1}{\sqrt{2\pi}}\exp{(-\frac{t^2}{2})}.
\end{equation}
With these preparation, the probability of each residual error $e_L(x_i)$ ($i=1,2,\ldots,N$) can be obtained by calculating $ \Phi(e_L(x_i))$. Then, the penalty weights $\theta_i$ ($i=1,2,\ldots,N$) can be assigned as
\begin{equation}\label{penalty_weights}
\theta_i=\Phi(e_L(x_i))=\frac{1}{\tau N}\sum_{k=1}^{N}K\left(\frac{\|e_L(x_i)-e_L(x_k)\|}{\tau}\right).
\end{equation}
With these penalty weights, the output weights $\beta^{*}=[\beta^{*}_1, \beta^{*}_2,\ldots,\beta^{*}_{L}]$ can be evaluated by solving the following WLS problem:
\begin{eqnarray}\label{wls2}
  \beta^{*}&=&\arg\min_{\beta}\:(H_L\beta-T)^{\mathrm{T}}\Lambda (H_L\beta-T)\nonumber\\
 &=&(H_L^{\mathrm{T}}\Lambda H_L)^{\dagger}H_L^{\mathrm{T}}\Lambda T,
\end{eqnarray}
where $\beta=[\beta_1, \beta_2,\ldots,\beta_{L}]$, $H_L=[h_1,h_2,\ldots,h_L]$, $\Lambda=\Theta^2=\mbox{diag}\{\theta_1,\theta_2,\ldots,\theta_N\}$.

In this paper, an alternating optimization (AO) strategy is applied for implementing RSCNs, which includes the process of building a SCN model with a set of suitable penalty weights that control the contribution of contaminated samples. The whole procedure begins with assigning equal penalty weights for all samples (i.e., $\theta_i=1$, and $\Lambda$ is an identity matrix) and building the SCN model, followed by updating these penalty weights according to (\ref{penalty_weights}), then repeating these two steps alternatively until some certain stopping criterion is reached. It should be clarified that the penalty weights are updated only when a round of the process of building the SCN model is completed.

Specifically, the penalty weights $\theta_i$ ($i=1,2,\ldots,N$) and the output weights $\beta$ can be calculated iteratively by applying the AO procedure, that is,
\begin{eqnarray}\label{theta_ite}
  \theta_i^{(\nu+1)}=\frac{1}{\tau N}\sum_{k=1}^{N}K\:\Bigg(\frac{e_L^{(\nu)}(x_i)-e_L^{(\nu)}(x_k)}{\tau}\Bigg)
\end{eqnarray}
and
\begin{equation}\label{beta_ite}
  \beta^{(\nu+1)}=(H_L^{\mathrm{T}}\Lambda^{(\nu+1)} H_L)^{\dagger}H_L^{\mathrm{T}}\Lambda^{(\nu+1)} T,
\end{equation}
where $\nu$ denotes the $\nu$-th iteration of the alternating optimization process, and $\Lambda^{(\nu+1)}=\mbox{diag}\{\theta_1^{(\nu+1)},\theta_2^{(\nu+1)},\ldots,\theta_N^{(\nu+1)}\}$. Here, we use $e_L^{(\nu)}(x_i)$ to represent the residual error value for $x_i$ with $\theta_i^{(\nu)}$ used as the present penalty weights in the RSCN model.

Distinguished from the original SCN framework that all training samples contribute equally to the objective function, our newly developed RSCNs treat individual samples differently and put more emphasis on data samples with higher reliability, which indeed corresponds to lager values of the penalty weights. That means if a training output $y_j$ is corrupted by outliers or noises, the sample pair ($x_j,y_j$) will provide less contribution to the cost function due to the relatively small value of its corresponding penalty weight. 

\textbf{Remark 2:} An important issue in design of RSCNs is about the termination criterion. As mentioned in Remark 1, the performance over a validation data set can be employed as a stopping condition. Unfortunately, this method does not make sense and cannot be applied for building RSCNs due to the presence of uncertainties  in the validation data. However, the validation testing criterion can still be used if a clean data set extracted from the true data distribution is available. In this work, we assume that a clean validation data set is ready to be used for this purpose. Our proposed RSC-KDE algorithm is summarized in the following pseudo code.

\begin{table}[htbp!]
\begin{center}
\small
%\begin{tabular}{lcl}
\begin{tabular}{p{0.92\columnwidth}}
\toprule
\textbf{RSC-KDE Algorithm } \\
\midrule
Given inputs $X\!=\!\{x_1,x_2,\ldots,x_N\}$, $x_i\in \mathds{R}^{d}$, outputs $T\!=\!\{t_1,t_2,\ldots,t_N\}$, $t_i\in \mathds{R}^{m}$; Set $L_{max}$ as the maximum number of hidden nodes, $\epsilon$ as the expected error tolerance, $P_{max}$ as the maximum times of random configuration, $I_{max}$ as the maximum number of alternating optimization; Choose a set of positive scalars $\Upsilon\!=\{\lambda_{min}\!:\!\Delta\lambda\!:\!\lambda_{max}\}$;\\
\midrule
\textbf{1.} Initialize $e_0\!:=[t_1,\ldots,t_N]^\mathrm{T}$, $0<r<1$, $\theta_i=1$, $\Lambda\!=\!\mbox{diag}\{\theta_1,\theta_2,\ldots,\theta_N\}$, $\Omega,W:=[\:\:]$;\\
\textbf{2.} \textbf{While} $\nu\leq I_{max}$ AND $\|e_0\|_{F}>\epsilon$, \textbf{Do}\\
\textbf{3.}\:\:\:\:\:\:\textbf{While} $L\leq L_{max}$ AND $\|e_0\|_{F}>\epsilon$, \textbf{Do}\\
\textbf{4.}\:\:\:\:\:\:\:\:\:\:\:\:\:\textbf{For} $\lambda \in \Upsilon$, \textbf{Do}\\
\textbf{5.}\:\:\:\:\:\:\:\:\:\:\:\:\:\:\:\:\:\textbf{For} $k=1,2\ldots,P_{max}$, \textbf{Do}\\
\textbf{6.}\:\:\:\:\:\:\:\:\:\:\:\:\:\:\:\:\:\:\:\:\:Randomly assign $\omega_L$ and $b_L$ from $[-\lambda,\lambda]^d$ and $[-\lambda,\lambda]$, respectively;\\
\textbf{7.}\:\:\:\:\:\:\:\:\:\:\:\:\:\:\:\:\:\:\:\:\:Calculate $\tilde{e}_{L-1}$, $\tilde{h}_L$, $\tilde{\xi}_{L,q}$ by  Eq. (\ref{factor1}), set $\mu_L=(1-r)/(L+1)$;\\
\textbf{8.}\:\:\:\:\:\:\:\:\:\:\:\:\:\:\:\:\:\:\:\:\:\textbf{If}\:\:\:$min\{\tilde{\xi}_{L,1},\tilde{\xi}_{L,2},...,\tilde{\xi}_{L,m}\}\geq 0$\\
\textbf{9.}\:\:\:\:\:\:\:\:\:\:\:\:\:\:\:\:\:\:\:\:\:\:\:\:\:\:\:\textbf{Save} $w_L$ and $b_L$ in $W$, $\tilde{\xi}_L$ in $\Omega$, respectively;\\
\textbf{10.}\:\:\:\:\:\:\:\:\:\:\:\:\:\:\:\:\:\:\:\:\:\textbf{Else} Go back to \textbf{Step 5}\\
\textbf{11.}\:\:\:\:\:\:\:\:\:\:\:\:\:\:\:\:\:\:\:\textbf{End If}\\
\textbf{12.}\:\:\:\:\:\:\:\:\:\:\:\:\:\:\:\textbf{End For}\:(corresponds to \textbf{Step 5}) \\
\textbf{13.}\:\:\:\:\:\:\:\:\:\:\:\textbf{If}\:\:$W$ is not empty\\
\textbf{14.}\:\:\:\:\:\:\:\:\:\:\:\:\:\:\:\:\:Find $w_L^{*}$, $b_L^{*}$ maximizing $\tilde{\xi}_L$ in $\Omega$, and set $H_L=[h^*_1,h^*_2,\ldots,h^*_L]$;\\
\textbf{15.}\:\:\:\:\:\:\:\:\:\:\:\:\:\:\:\:\:\textbf{Break} (go to \textbf{Step 19}); \\
\textbf{16.}\:\:\:\:\:\:\:\:\:\:\:\textbf{Else} Randomly take $\tau\in (0,1-r)$, renew $r:=r+\tau$, return to \textbf{Step 5};\\
\textbf{17.}\:\:\:\:\:\:\:\:\:\:\:\textbf{End If}\\
\textbf{18.}\:\:\:\:\:\:\:\:\:\:\:\textbf{End For} (corresponds to \textbf{Step 4})\\
\textbf{19.}\:\:\:\:\:\:\:\:\:\:\:Calculate $\beta^{*}=[\beta^{*}_1,\beta^{*}_2,\ldots,\beta^{*}_L]$ based on Eq. (\ref{beta_ite});\\
\textbf{20.}\:\:\:\:\:\:\:\:\:\:\:Calculate $e_L=H_L\beta^{*}-T$ and obtain $\tilde{e}_L$;\\
\textbf{21.}\:\:\:\:\:\:\:\:\:\:\:Renew ${e}_0:=\tilde{e}_L$, $L:=L+1$; \\
\textbf{22.}\:\:\:\:\textbf{End While}\\
\textbf{23.}\:\:\:\:Update $\theta_i$ by Eq. (\ref{theta_ite}), renew $\Lambda=\mbox{diag}\{\theta_1,\theta_2,\ldots,\theta_N\}$ and $\nu:=\nu+1$;\\
\textbf{24.} \textbf{End While}\\
\textbf{25.} \textbf{Return} $\Lambda=\mbox{diag}\{\theta_1,\ldots,\theta_N\}$, $\beta^{*}=[\beta^{*}_1,\beta^{*}_2,\ldots,\beta^{*}_L]$, $\omega^{*}=[\omega^{*}_1,\omega^{*}_2,\ldots,\omega^{*}_L]$, $b^{*}=[b^{*}_1,b^{*}_2,\ldots,b^{*}_L]$.\\
\bottomrule
\end{tabular}
\end{center}
\end{table}

\section{Performance Evaluation}
This section reports some simulation results on a function approximation, four benchmark datasets from
KEEL \!\!\!\! \footnote{KEEL: http://www.keel.es/}\!, and an industrial application \cite{Dai2015}. The proposed RSC-KDE algorithm is compared to other three randomized algorithms: RVFL \cite{Igelnik1995}, improved RVFL \cite{Dai2015}, and the probabilistic learning algorithm PRNNRW \cite{Cao2015}. All comparisons are conducted under several scenarios with different system settings on learning parameters and noise levels. The Root Mean Squared Error (RMSE) is used to evaluate the generalization capability of each algorithm over the outlier-free test datasets. In addition, a robustness analysis on the setting of $\nu$ and $L_{max}$ is given for the case study.

The input and output values are normalized into [0,1] before artificially adding certain level of outliers. The maximum times of random configuration $T_{max}$ in RSC-KDE is set as 100, and the sigmoidal activation function $g(x)=1/(1+\exp(-x))$ is used in all simulations.

\subsection{Function Approximation}
Consider the following function approximation problem \cite{Tyukin2009}:
\begin{equation*}
y = 0.2e^{-(10x-4)^{2}}+0.5e^{-(80x-40)^{2}}+0.3e^{-(80x-20)^{2}},\:\:x\in[-1,1].
\end{equation*}
The training dataset contains 600 points which are randomly generated from the uniform distribution [-1,1]. The test dataset, of size 600, is generated from a regularly spaced grid on [-1,1]. We purposely introduce outliers into the training dataset: A variable percentage $\xi$ of the data points is selected randomly and their corresponding function values (y) are substituted by background noises with values uniformly distributed over [-0.2, 0.8]. To show the advantage of RSC-KDE in uncertain data modelling, we make comparisons on the performance among these four algorithms at each outlier percentage, i.e., $\zeta=\{0\%,5\%,10\%,15\%,20\%,25\%,30\%\}$.
\begin{figure}[htbp]
\centering
\subfigure[]{\includegraphics[width=0.45\textwidth]{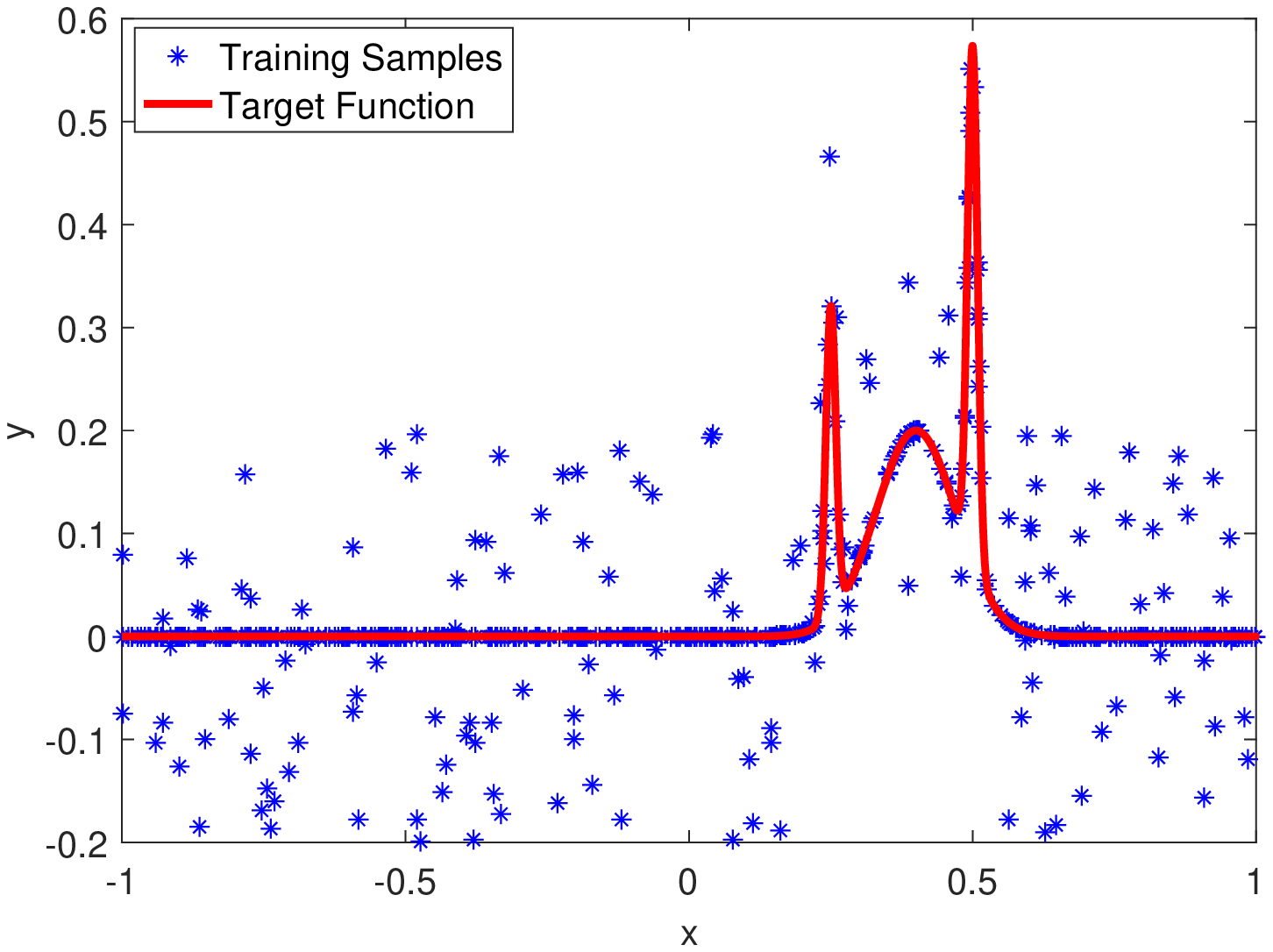}}
\subfigure[]{\includegraphics[width=0.45\textwidth]{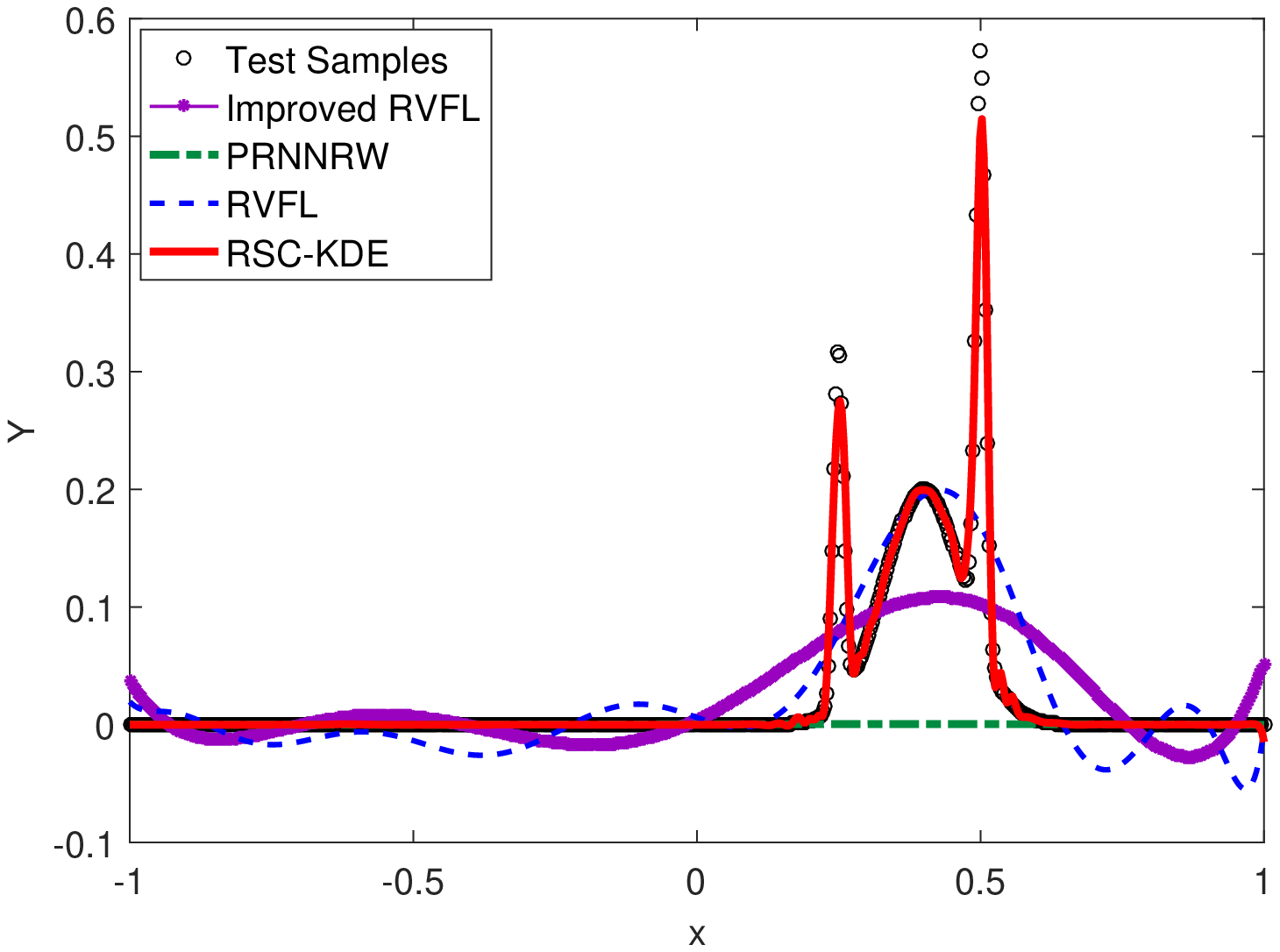}}
\caption{(a) 600 training samples used for function approximation at $\zeta=25\%$, along with target function shown in red line; (b) Approximation performance on the test dataset by four learning algorithms at $\zeta=25\%$.}\label{fig:1}
\end{figure}

Figure 1 (a) depicts the training samples with the outlier percentage at $\xi=25\%$, Figure 1 (b) shows the learners' performance on the test dataset from the four algorithms, in which the proposed RSC-KDE exhibits the best performance compared with the other three methods. For RVFL, Improved RVFL and PRNNRW, we examined different scope settings for the random parameters, i.e., $w,b\in [-\lambda,\lambda]$, $\lambda=1,10,30,50,100,150$, to demonstrate its significant impact on the randomized learners' performance. For each $\lambda$, different network architectures (e.g. $L=40,60,100,120,150,200$) are used to find the pair ($\lambda,L$) leading to the most favorable performance. For $\lambda=1,30,50,100$, we demonstrate the test results of the four algorithms in Figure 2, where the average errors and standard deviations of RMSE (based on 100 trials) are plotted for each outlier percentage. It is clear that our proposed RSC-KDE algorithm outperforms the other methods for each case. In particular, for $\lambda=1$, the approximation performance of all other three algorithms are far worse than an acceptable level. Obviously, if the scope setting is improper, the randomized learner models can not be expected to perform at all, as reported in \cite{LiandWang2016,WangandLi_SCN}. As the outlier percentage becomes very low, RSCN outperforms other randomized learner models, which aligns well with the consequence reported in \cite{WangandLi_SCN}. Given a relatively higher outlier percentage, for example $\zeta=30\%$, RSC-KDE produces a promising result with RMSE of $0.0138\pm0.0027$, that is much better than the other three algorithms. More results for $\lambda=1,30,50,100$ at $\zeta=10\%,15\%,20\%,25\%$ are reported in Table 1.

\begin{figure}[htbp!]
\centering
\subfigure[$\lambda=1$]{\includegraphics[width=0.45\textwidth]{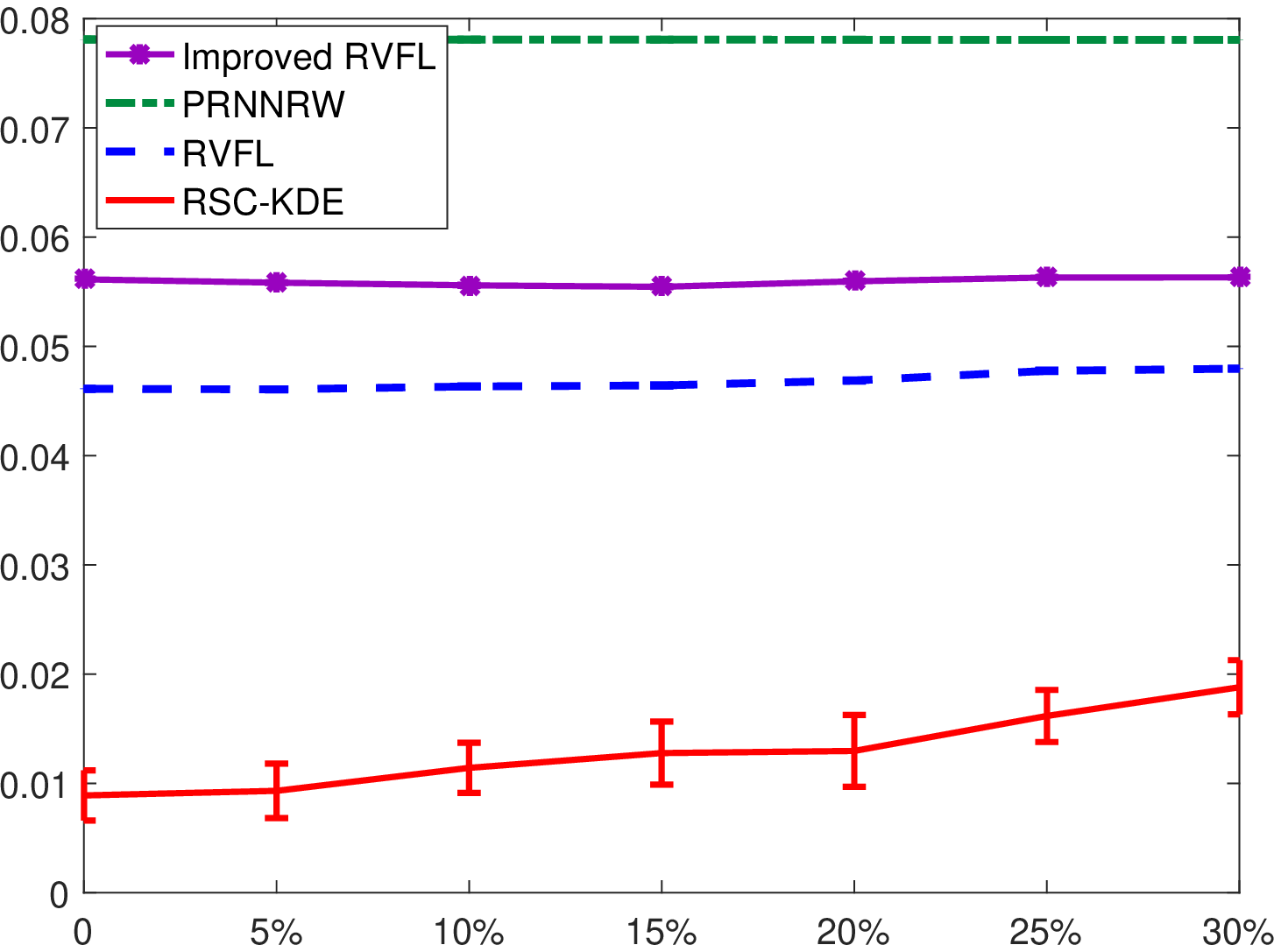}}
\subfigure[$\lambda=30$]{\includegraphics[width=0.45\textwidth]{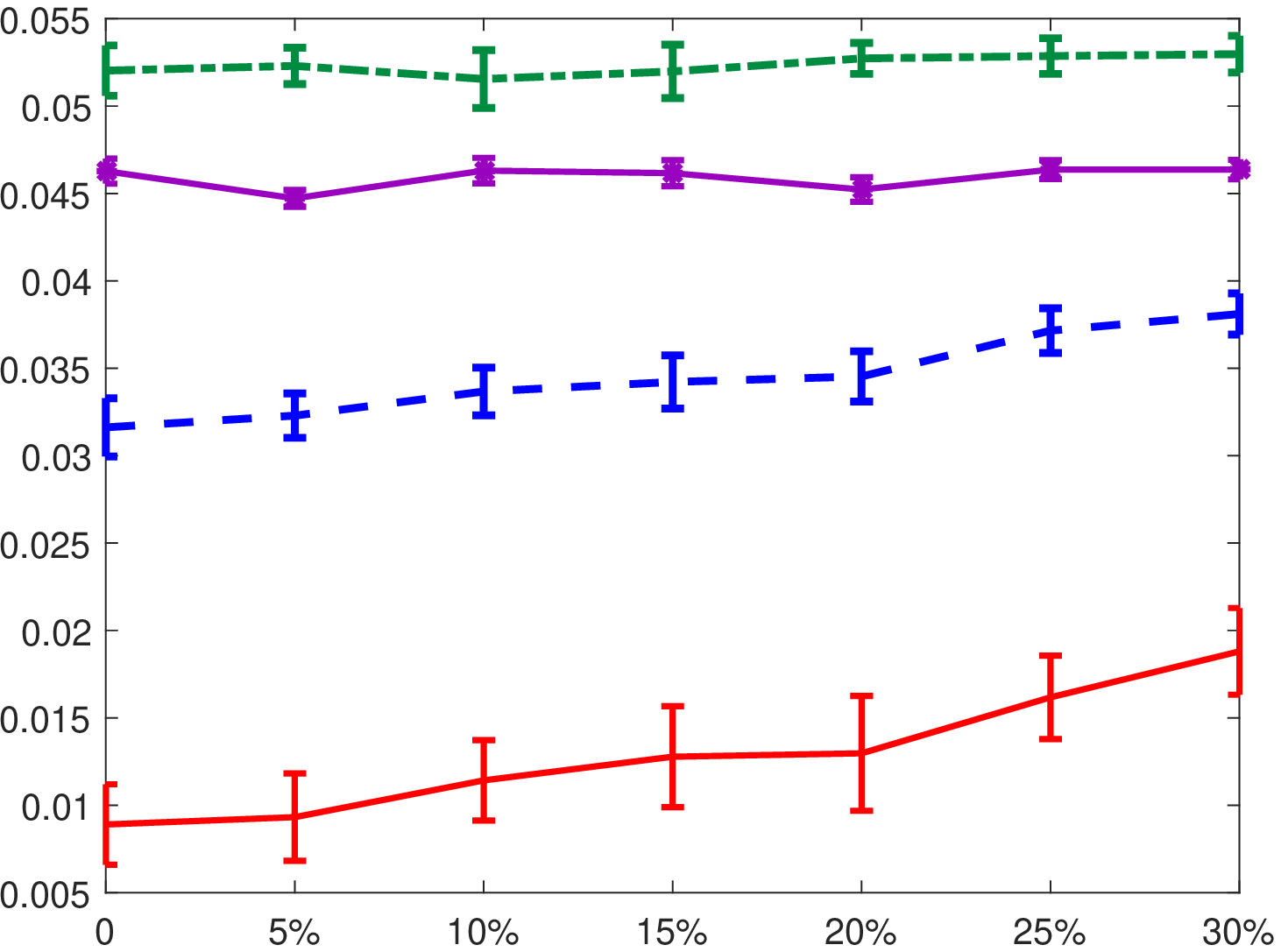}}
\subfigure[$\lambda=50$]{\includegraphics[width=0.45\textwidth]{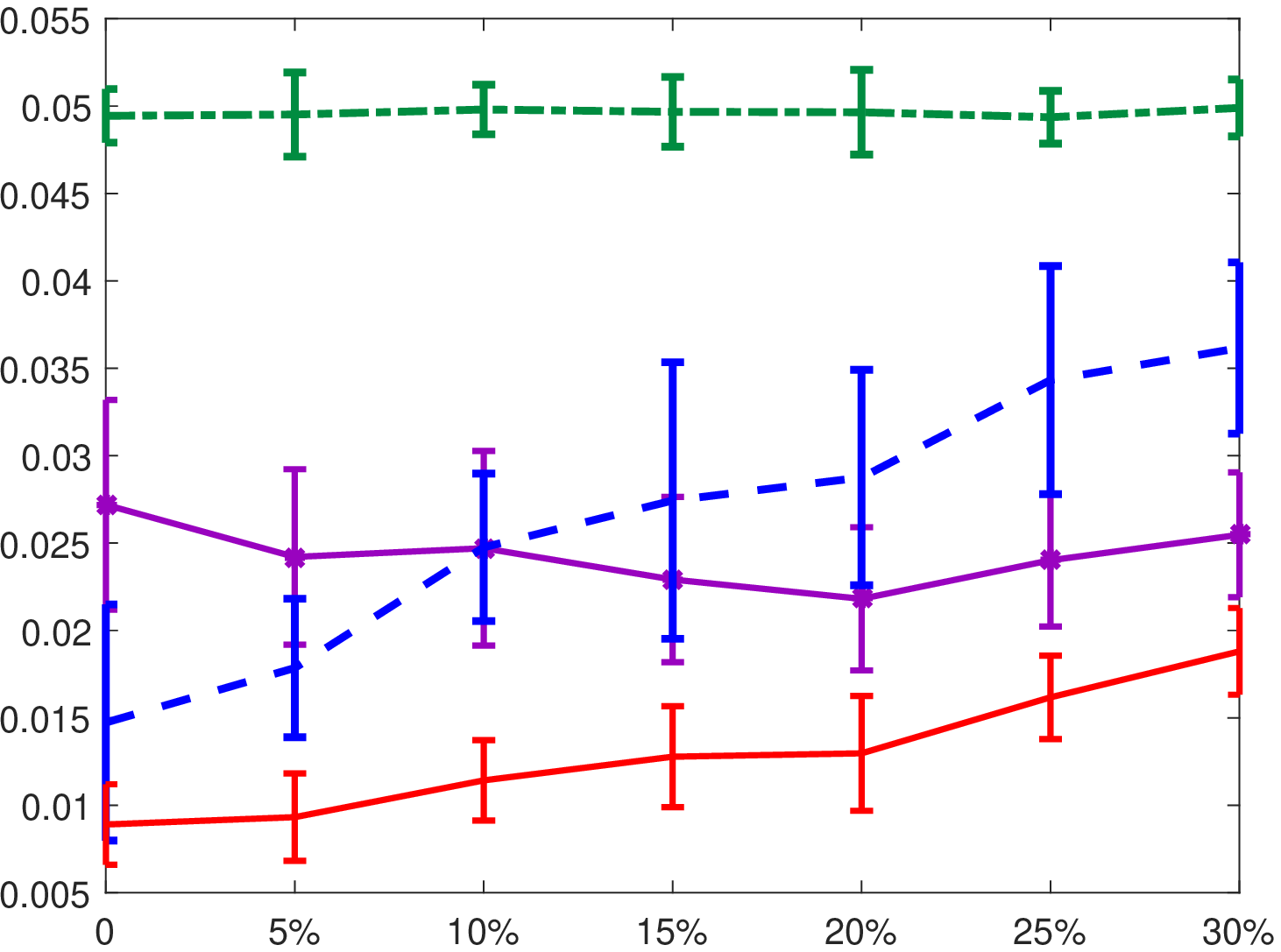}}
\subfigure[$\lambda=100$]{\includegraphics[width=0.45\textwidth]{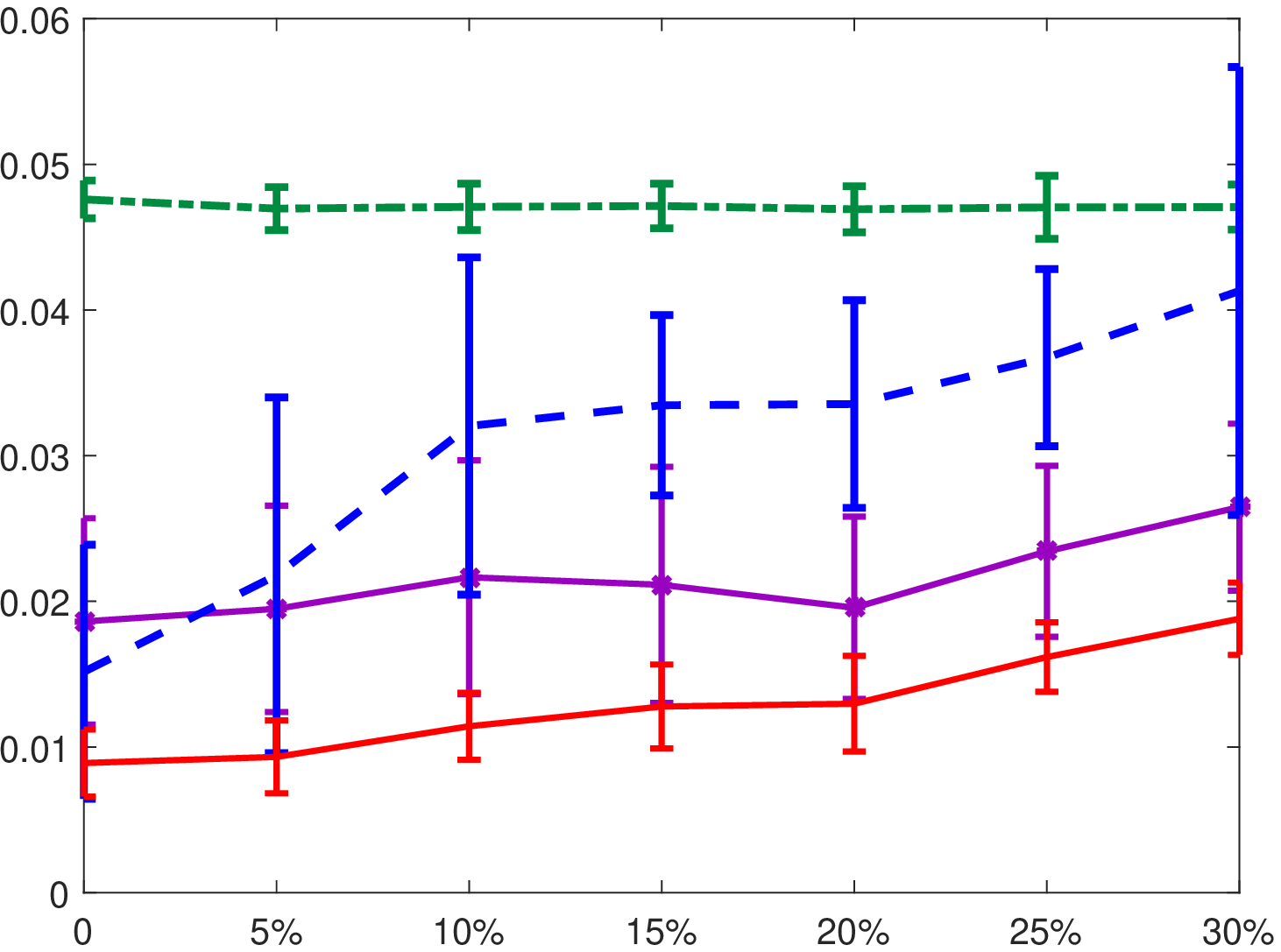}}
\caption{Test RMSE comparison on the function approximation among the four algorithms with different outlier percentage $\zeta$. $\lambda=1,10,50,100$ and $L=100$ are used in RVFL, Improved RVFL and PRNNRW.}\label{fig:2}
\end{figure}

\begin{table}[h!]\label{table1}
\centering
\footnotesize
{\caption{Performance comparisons on the function approximation}}
\begin{center}
\begin{tabular}{ccccccc}\hline
\multirow{2}*{Scope Setting} &\multirow{2}*{Algorithm}&\multicolumn{4}{c}{Test Performance at Different Outlier Percentage (MEAN, STD)} \\
\cline{3-6}
 & &10$\%$&15$\%$&20$\%$&25$\%$\\
\hline
\multirow{4}*{$\lambda=1$}&   RVFL  &0.0463,2.0e-05 &0.0464,2.0e-05  &0.0469,2.0e-05 &0.0478,2.0e-05\\
&Improved RVFL     &0.0556,5.1e-05 &0.0555,3.0e-05  &0.0560,4.0e-05 &0.0563,3.3e-05 \\
&PRNNRW     &0.0781,\textbf{6.6e-06} &0.0781,\textbf{7.3e-06}  &0.0781,\textbf{7.3e-06} &0.0781,\textbf{6.8e-06} \\
&RSC-KDE    &\textbf{0.0090},0.0016 &\textbf{0.0098},0.0022  &\textbf{0.0104},0.0035 &\textbf{0.0117},0.0022 \\
\hline
\multirow{4}*{$\lambda=30$}&   RVFL  &0.0225,0.0037 &0.0249,0.0039  &0.0264,0.0033 &0.0326,0.0032 \\
&Improved RVFL     &0.0344,0.0030 &0.0326,0.0036  &0.0319,0.0034 &0.0322,0.0028 \\
&PRNNRW     &0.0491,0.0029 &0.0490,\textbf{0.0013} &0.0490,\textbf{0.0017} &0.0492,\textbf{0.0010} \\
&RSC-KDE     &\textbf{0.0090},\textbf{0.0016} &\textbf{0.0098},0.0022  &\textbf{0.0104},0.0035 &\textbf{0.0117},0.0022 \\
\hline
\multirow{4}*{$\lambda=50$}&   RVFL  &0.0248,0.0042&0.0274,0.0079 &0.0288,0.0062  &0.0343,0.0065 \\
&Improved RVFL &0.0247,0.0056 &0.0229,0.0047 &0.0218,0.0041 &0.0240,0.0038 \\
&PRNNRW     &0.0498,\textbf{0.0014} &0.0497,\textbf{0.0020}  &0.0496,\textbf{0.0024} &0.0494,\textbf{0.0015}\\
&RSC-KDE     &\textbf{0.0090},0.0016 &\textbf{0.0098},0.0022  &\textbf{0.0104},0.0035 &\textbf{0.0117},0.0022 \\
\hline
\multirow{4}*{$\lambda=150$}&   RVFL  &0.0345,0.0166 &0.0353,0.0062 &0.0380,0.0099  &0.0386,0.0086 \\
&Improved RVFL     &0.0229,0.0082 &0.0214,0.0074  &0.0228,0.0084 &0.0284,0.0076 \\
&PRNNRW     &0.0445,0.0031 &0.0439,0.0032  &0.0438,\textbf{0.0030} &0.0441,0.0037 \\
&RSC-KDE     &\textbf{0.0090},\textbf{0.0016} &\textbf{0.0098},\textbf{0.0022}  &\textbf{0.0104},0.0035 &\textbf{0.0117},\textbf{0.0022} \\
\hline
\end{tabular}
\end{center}
\end{table}

%%------------------------------------------------------Benchmarkdatasets---------------------------------------------
\subsection{Benchmark Datasets}
 Table 2 gives some statistics on the four datasets used in our simulations. We  randomly chose 75$\%$ of the whole samples as the training dataset while take the remainders as the test dataset. A similar strategy as done in the function approximation problem is applied to introduce different percentages of outliers into the training dataset. That is, for each normalized training dataset, a variable percentage $\zeta$ of the data points are selected at random and the associated output values are substituted with background noises that are uniformly distributed on the range [-0.5,0.5]. Finally, the contaminated output values are distributed over [-0.5,1.5] instead of [0,1], while the test dataset is outlier-free for the assessment purpose. 
\begin{table}[htbp!]\label{table2}
\caption{Statistics of the benchmark datasets}
\begin{center}
\begin{tabular}{ccccc}
\toprule
No.&Name&   Instances& Features \\
\midrule
1& stock& 950& 9 \\
2& laser& 993& 4 \\
3& concrete& 1030& 8 \\
4& treasury& 1049& 15  \\
\bottomrule
\end{tabular}
\end{center}
\end{table}

For each benchmark dataset, we evaluate the performance of RVFL, Improved RVFL and PRNNRW with different settings of $\lambda$ and $L$, for example, $\lambda=\{0.1,0.5,1,3,5\}$ and $L=\{30,50,100,150,200\}$, respectively. We conduct 50 independent trials for each case ($\lambda$ and $L$) and calculate their mean values and standard deviations of RMSE at different percentage of outliers. In Figure 3, we plot the comparison results for RVFL, Improved RVFL, and PRNNRW with $\lambda=1$, and it shows that our proposed RSC-KDE algorithm outperforms the others at each outlier percentage. From the results reported in Table 3, it is observed that our proposed RSC-KDE algorithm outperform the others for all these four datasets at each outlier percentage, despite that the results obtained by other three algorithms are the `best' ones selected from all results with  various settings of $\lambda$ and $L$. Specifically in Figure 3, PRNNRW (with $\lambda=1$) exhibits the worst accuracy on stock, laser, concrete, but performs better than RVFL and Improved RVFL on treasury. Also in Table 3, the results from PRNNRW, as obtained by the most appropriate combination of $\lambda$ and $L$ from the set $\{0.1, 0.5, 1, 3, 5\}$ and $\{30, 50, 100, 150, 200\}$, respectively, are much better than that shown in Figure 3.  Indeed, for RVFL, Improved RVFL and PRNNRW, a common practice to determine a suitable scope setting and a reasonable network architecture is based on the trial-and-error method, while the proposed RSC-KDE works robustly with much less human intervention on the parameter setting.

\begin{figure*}[h!]
\centering
\subfigure[stock]{\includegraphics[width=0.45\textwidth]{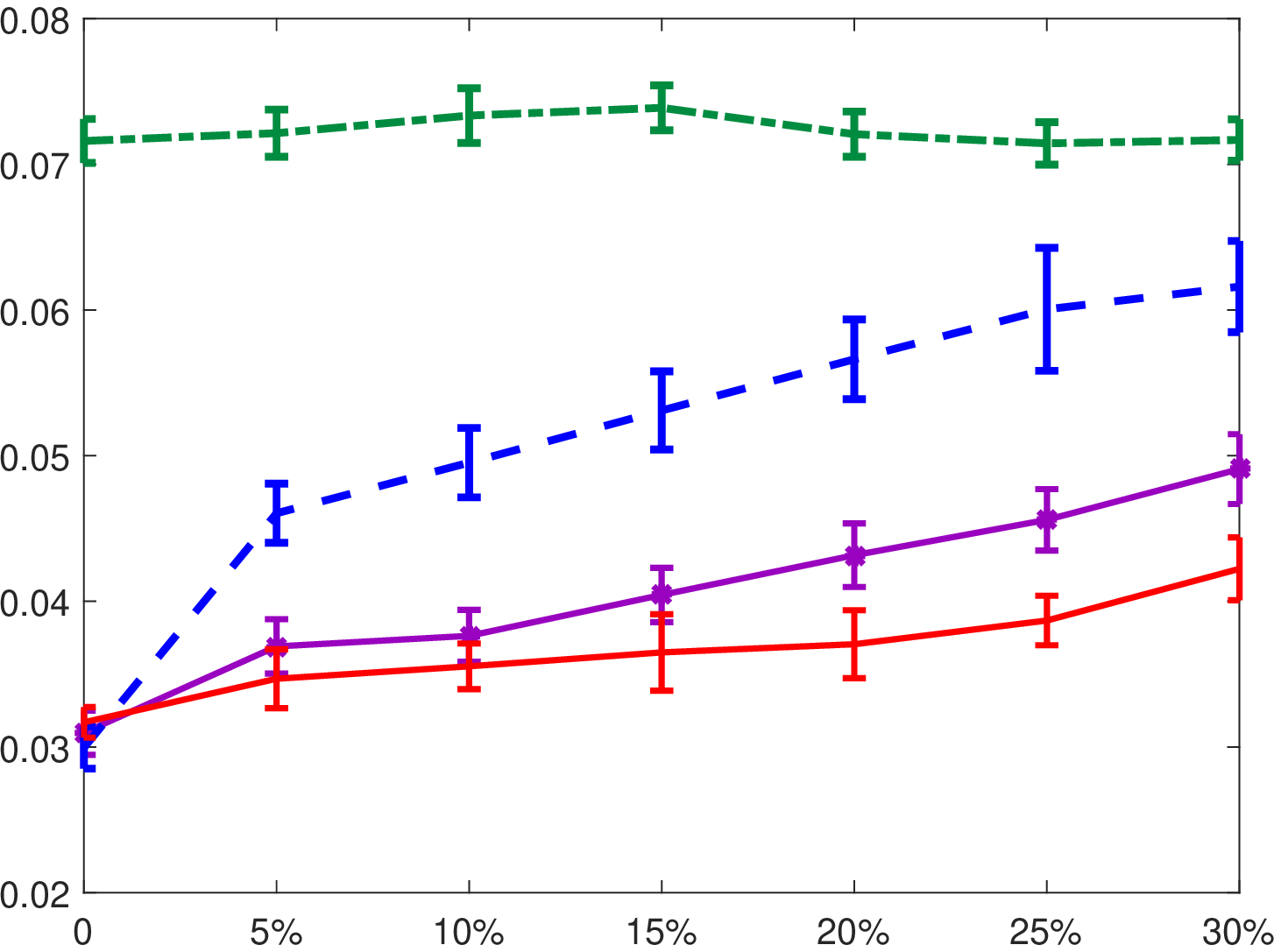}}
\subfigure[laser]{\includegraphics[width=0.45\textwidth]{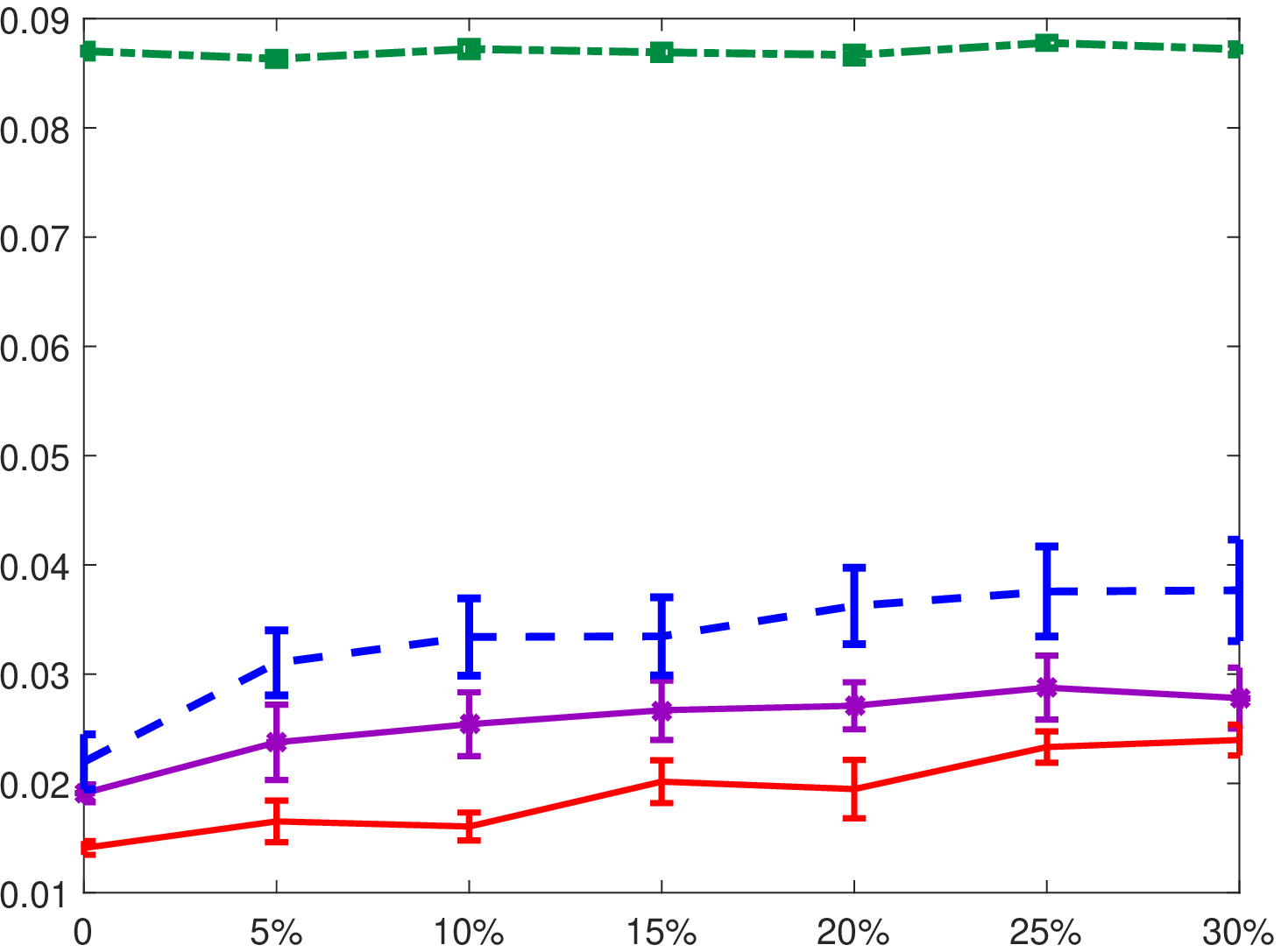}}
\subfigure[concrete]{\includegraphics[width=0.45\textwidth]{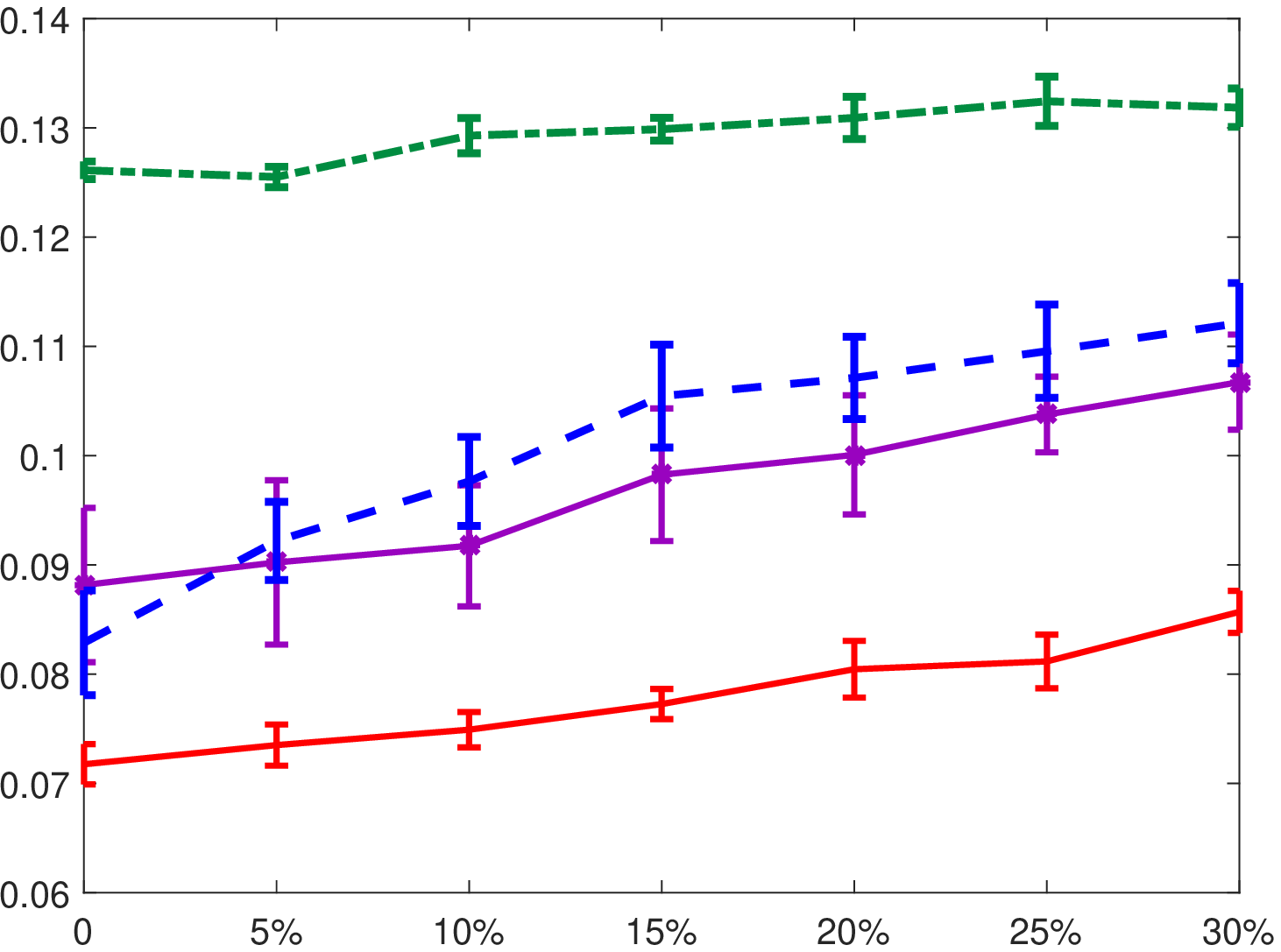}}
\subfigure[treasury]{\includegraphics[width=0.45\textwidth]{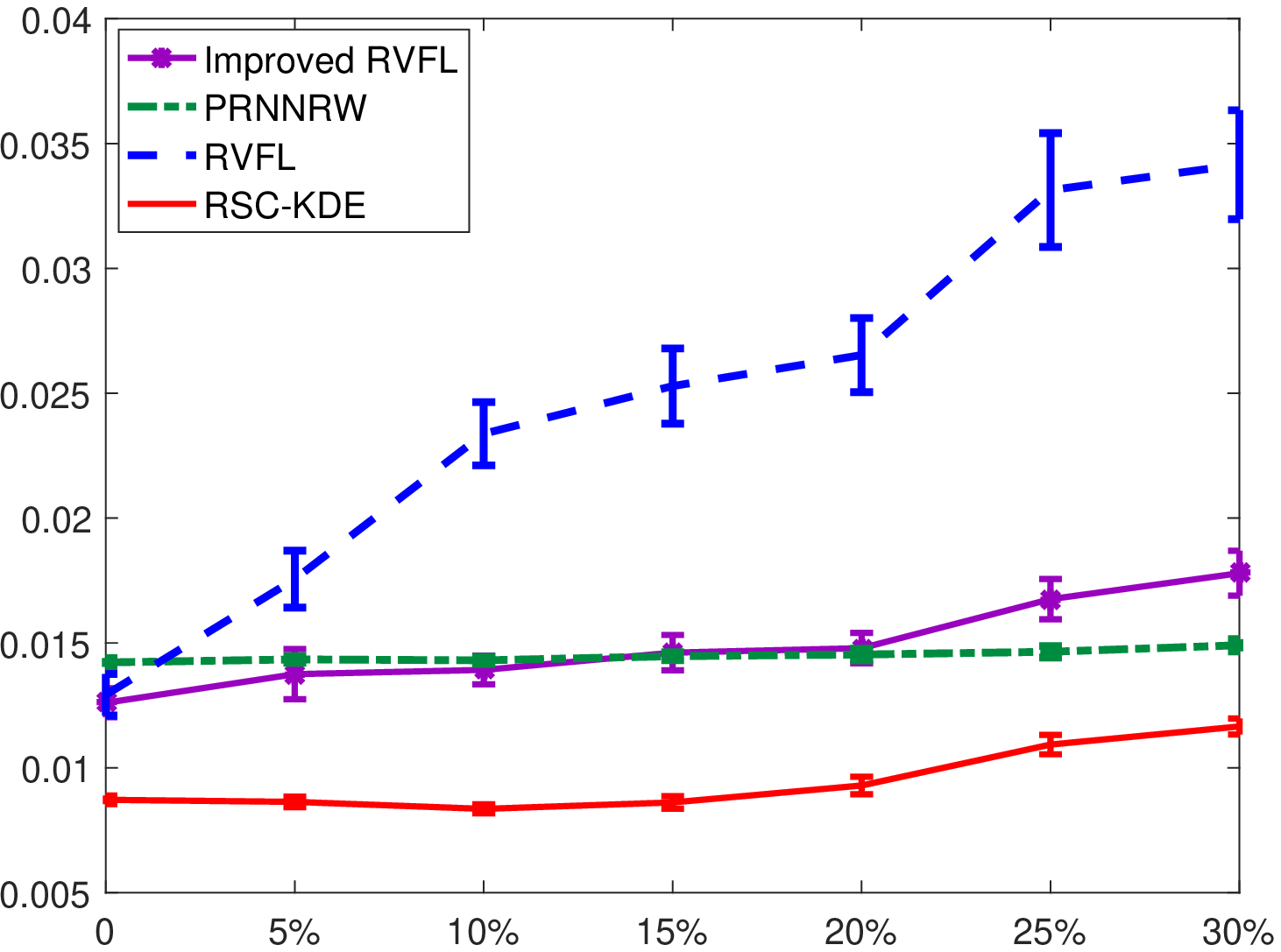}}
\caption{Test RMSE comparison on four benchmark datasets among four algorithms with different outlier percentage $\zeta$. $\lambda=1$  and $L=150$ are used in RVFL, Improved RVFL, and PRNNRW.}\label{fig:3}
\end{figure*}

\begin{table*}[htbp!]
\centering
\footnotesize
{\caption{Performance comparisons on the benchmark datasets}\label{tab:3}}
\begin{center}
\begin{tabular}{ccccccc}\hline
\multirow{2}*{Dataset} &\multirow{2}*{Algorithm}&\multicolumn{4}{c}{Test Performance at Different Outlier Percentage (MEAN,STD)} \\
\cline{3-6}
 && 10$\%$ & 15$\%$ & 20$\%$ & 25$\%$ \\
\hline
\multirow{4}*{stock}&   RVFL  &0.0495,0.0024 & 0.0531,0.0027 &0.0554,0.0032 &0.0590,0.0036 \\
                    &Improved RVFL  &0.0373,0.0023 &0.0404,0.0022 &0.0425,0.0014 &0.0456,0.0023 \\
                    &PRNNRW &0.0378,0.0014 &0.0387,\textbf{0.0014} &0.0388,\textbf{0.0011} &0.0392,0.0017 \\
                     &RSC-KDE   &\textbf{0.0317},\textbf{0.0014} &\textbf{0.0322},0.0016 &\textbf{0.0328},0.0012 &\textbf{0.0342},\textbf{0.0012} \\
\hline
\multirow{4}*{laser}&   RVFL  &0.0318,0.0033 &0.0323,0.0029 &0.0343,0.0030 &0.0359,0.0038 \\
&Improved RVFL        &0.0239,0.0023 &0.0260,0.0024 &0.0264,0.0021 &0.0277,0.0039 
\\
&PRNNRW       &0.0424,0.0022&0.0424,0.0024 &0.0421,\textbf{0.0026} &0.0428,0.0027 \\
&RSC-KDE  &\textbf{0.0161},\textbf{0.0013} &\textbf{0.0202},\textbf{0.0020} &\textbf{0.0195},0.0027 &\textbf{0.0233},\textbf{0.0014} 
\\
\hline
\multirow{4}*{concrete}&   RVFL &0.0975,0.0039 &0.1038,0.0047 &0.1068,0.0047 &0.1092,0.0037 
 \\
&Improved RVFL        &0.0903,0.0070 &0.0967,0.0064 &0.0991,0.0043 &0.1022,0.0032 
 \\
&PRNNRW        &0.1019,0.0029 &0.1008,0.0030 &0.1013,\textbf{0.0025} &0.1034,0.0031 \\
&RSC-KDE   &\textbf{0.0749},\textbf{0.0016} &\textbf{0.0773},\textbf{0.0014} &\textbf{0.0805},0.0026 &\textbf{0.0812},\textbf{0.0025} \\
\hline
\multirow{4}*{treasury}&   RVFL    &0.0231,0.0012 &0.0253,0.0011 &0.0265,0.0013 &0.0325,0.0016  \\
&Improved RVFL       &0.0135,0.0005 &0.0145,0.0004 &0.0145,0.0005 &0.0166,0.0005 \\
&PRNNRW      &0.0130,0.0004 &0.0130,0.0004 &0.0131,\textbf{0.0002} &0.0131,\textbf{0.0002} \\
&RSC-KDE   &\textbf{0.0084},\textbf{0.0002} &\textbf{0.0086},\textbf{0.0002}&\textbf{0.0093},0.0003&\textbf{0.0109},0.0004\\
\hline
\end{tabular}
\end{center}
\end{table*}
%%---------------------------------------------------PSE real application------------------------------------------
\subsection{Particle Size Estimation of Mineral Grinding Process: A Case Study}
In this section, we make a further investigation on the merits of our proposed RSC-KDE algorithm by using a dataset from process industry  \cite{Dai2015}. Figure 4 depicts the grinding process where the coarse fresh ore $O_F$ is fed into the ball mill through the conveyor. Meanwhile, a certain amount of mill water $Q_F$ is added through a pipe to maintain a proper pulp density. At this stage, the steel balls within the mill crush the coarse ore to a finer size alone with the knocking and tumbling actions. After  grinding, the mixed ore pulp that includes both coarser and finer particles is discharged continuously from the mill into the spiral selector for further classification with assistance of dilution water $Q_D$ mixed to the ore pulp. Next, the pulp is separated into the overflow and underflow pulp. Finally, the underflow pulp with coarser particles is recycled back to the mill for re-grinding, whilst the overflow pulp with finer particles (as product) is further proceeded. As can be seen that the particle size estimation plays an important role in this circling procedure. 

Particle size estimation of mineral grinding process can be formulated as a regression problem, with three input variables including the fresh ore feed rate $O_F$, the mill water flow rate $Q_F$, and the dilution water flow rate $Q_D$, and with a single output of the unmodelled dynamics, namely $\triangle r$.  Denoted by $x=[O_F,Q_F,Q_D]^{T}$ and $y$ as the input vector and the output (refers to the estimation of the unmodelled dynamics $\triangle \tilde{r}$), respectively. Let $\tilde{r}$ represent an estimate of the particle size from a mathematical (mechanism) model. Thus,  the final estimated value of the particle size $\tilde{r}$ can be evaluated by  $\tilde{r}:=\tilde{r}+\triangle \tilde{r}$. 
\begin{figure*}[htbp]
\centering
\includegraphics[width=11.6cm]{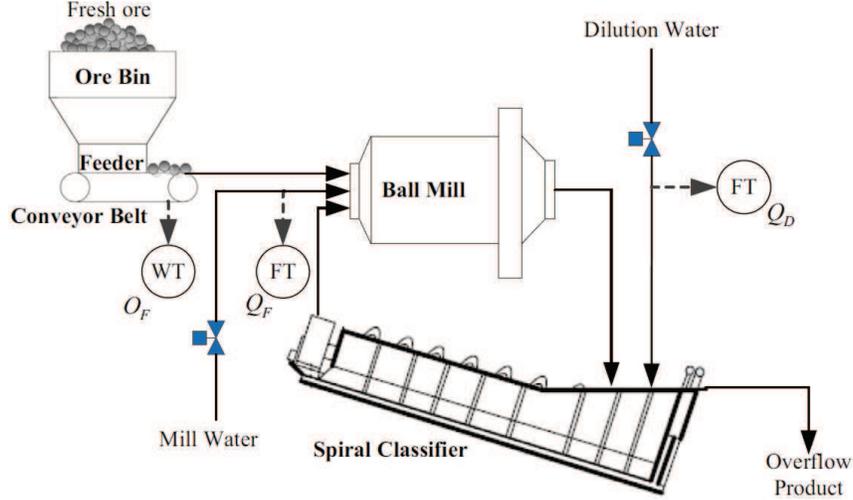}
\caption{Flow chart of mineral grinding process \cite{Dai2015}.}
\end{figure*}\label{fig:4}

In this case study, 300 training samples and 300 test samples were collected from a hardware-in-the-loop (HIL) platform \cite{Dai2015}, which is composed of  the following five subsystems: an optimal setting control subsystem, a human supervision subsystem, a DCS control subsystem, a virtual actuator and sensors subsystem, and a virtual operation process subsystem. For detailed descriptions on the operational functionalities of these subsystems, readers can refer to \cite{Dai2015}. Both the input and output values are normalized into [0,1]. Then, different levels of outliers are added into the normalized training dataset in the similar way as done  in the previous simulations, i.e, a variable percentage $\zeta$ of data points are selected randomly and the corresponding output values are corrupted by background noises followed the uniform distribution [-0.5,0.5]. As a result, the output values are distributed in the range [-0.5,1.5], while the test samples are outlier-free. 

The performance of these four algorithms are evaluated at several outlier percentages, i.e., $\zeta=\{0\%,5\%,10\%,15\%,20\%,25\%,30\%\}$. Specifically, for RVFL, Improved RVFL, and PRNNRW, different settings of $\lambda$ and $L$ are used in this study. Each comparison is based on 50 independent trials, the mean value and standard deviation of RMSE are recorded at each outlier percentage. The test performance of the four algorithms are depicted in Figure 5, where the presented results for RVFL, Improved RVFL and PRNNRW with each scope setting ($\lambda=0.1,0.5,1,5$) correspond to the `best' records among the trails using different $L$ ($L=10,30,50,100,150$). It can be easily found that our proposed RSC-KDE algorithm outperforms the other three methods in most cases. Although the Improved RVFLN exhibits very close performance when the outlier percentage is relatively lower, RSC-KDE algorithm has demonstrated the best at robustness even at high outlier contamination rate. When $\lambda=0.5$ and $\lambda=1$, the performance of PRNNRW has been improved a lot compared with $\lambda=0.1$, but are still unacceptable in comparison with Improved RVFL and RSC-KDE. For both RVFL and Improved RVFLN, there is no remarkable  difference between the results with $\lambda=0.5$ and $\lambda=1$. Interestingly, Figure 5 (d) shows that results from PRNNRW (with $\lambda=5$) at relatively higher outlier percentages (e.g. $\zeta=20\%, 25\%, 30\%$) are slightly better than RSC-KDE. However, this result needs a suitable scope setting (i.e. $\lambda=5$), which is time-consuming in practice. In contrast, the proposed RSC-KDE can lead to good performance than the others without any user-oriented trials for parameter setting. Table 4 summarizes the records of our proposed RSC-KDE algorithm as well as the `best' results obtained from RVFL, Improved RVFL and PRNNRW with the `most appropriate' parameter setting on $L$ for each $\lambda$. The impact of the scope setting for the other three randomized algorithms can be seen through comparing their records at each outlier percentage. Specifically for $\lambda=1$, the test results from the four algorithms are shown in Figure 6, where both the real and estimated values of the normalized particle size (of the whole test samples) are plotted. 
\begin{figure}[!h]
\centering
\subfigure[$\lambda=0.1$]{\includegraphics[width=0.45\textwidth]{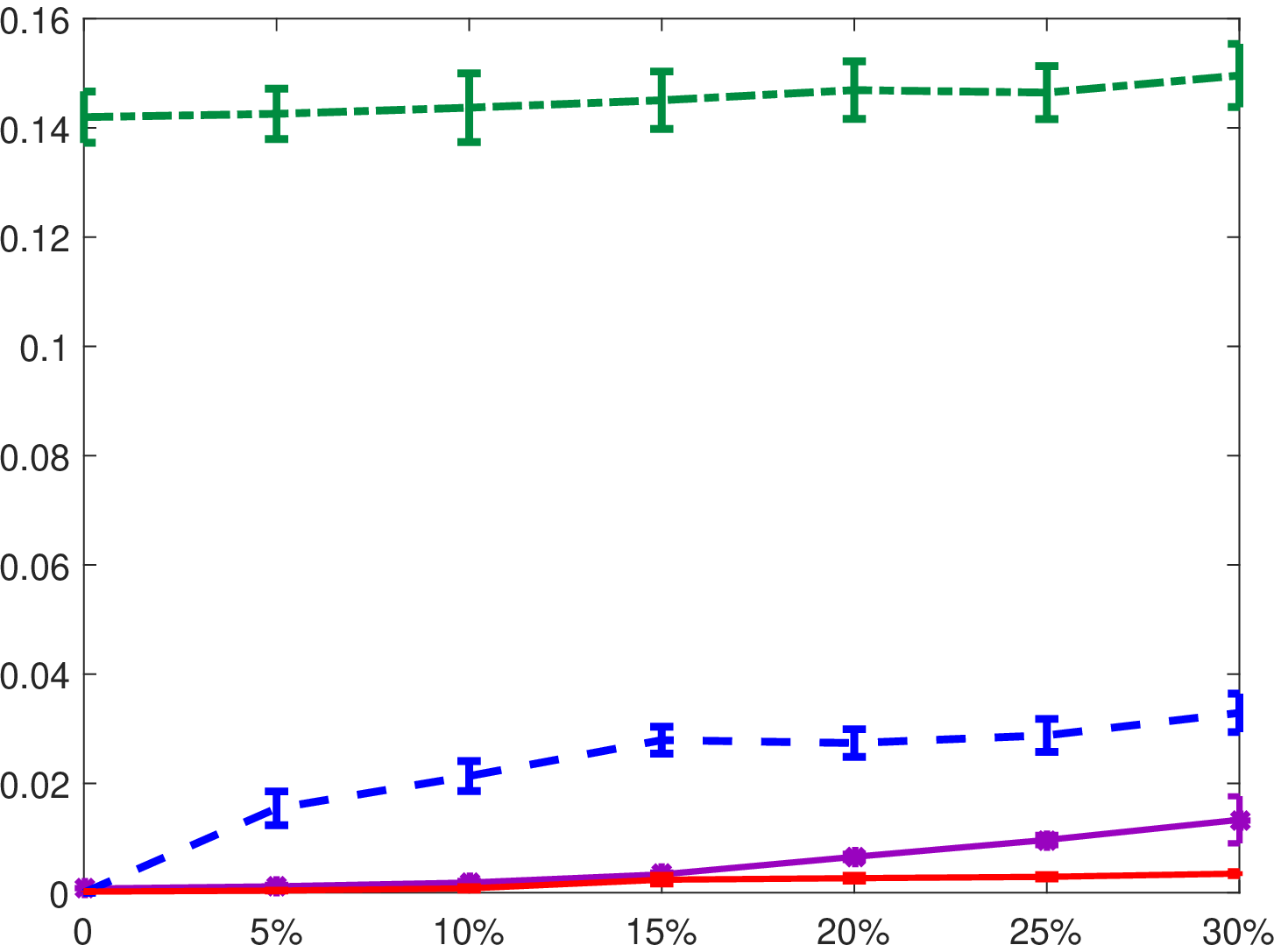}}
\subfigure[$\lambda=0.5$]{\includegraphics[width=0.45\textwidth]{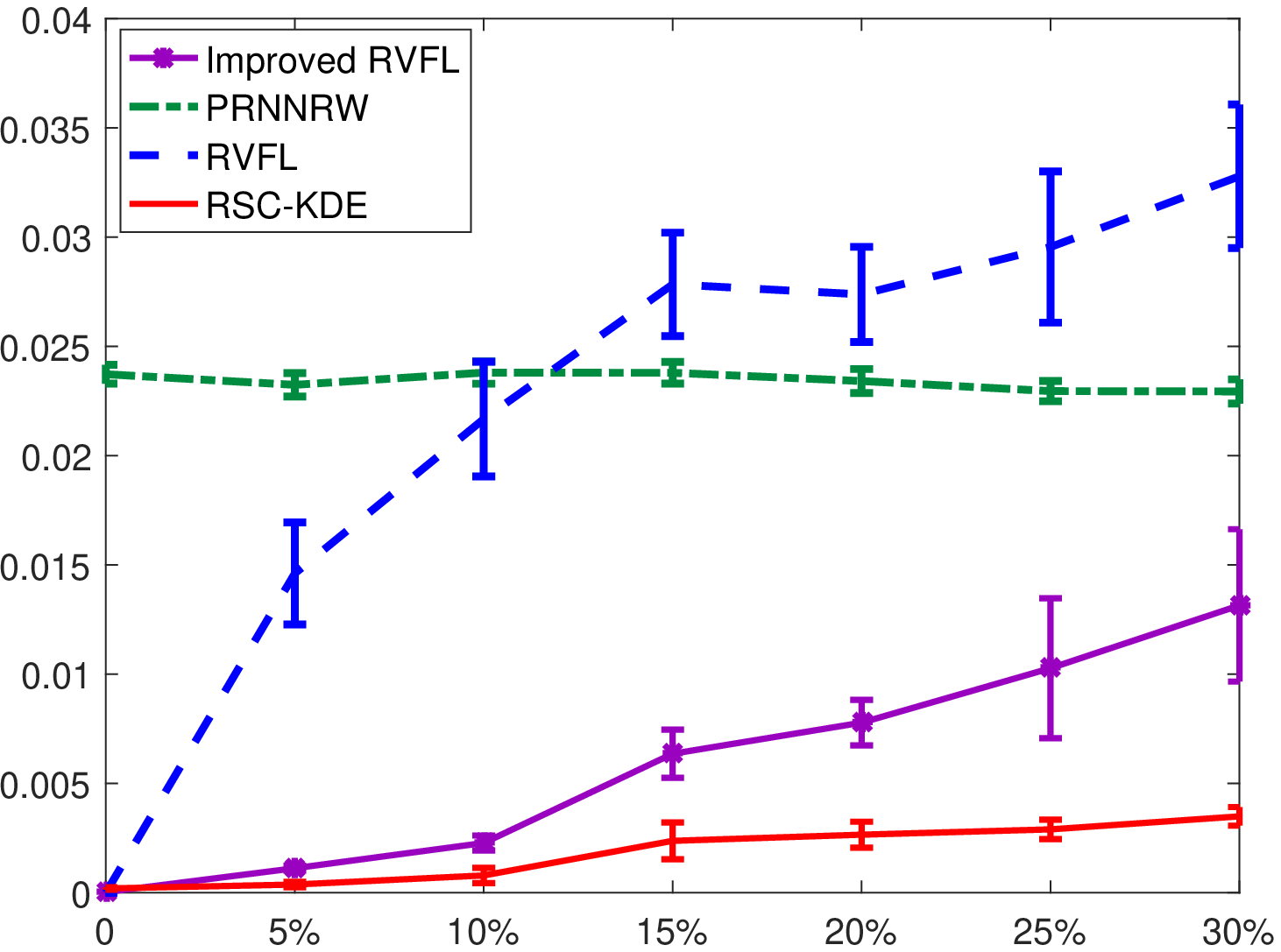}}
\subfigure[$\lambda=1$]{\includegraphics[width=0.45\textwidth]{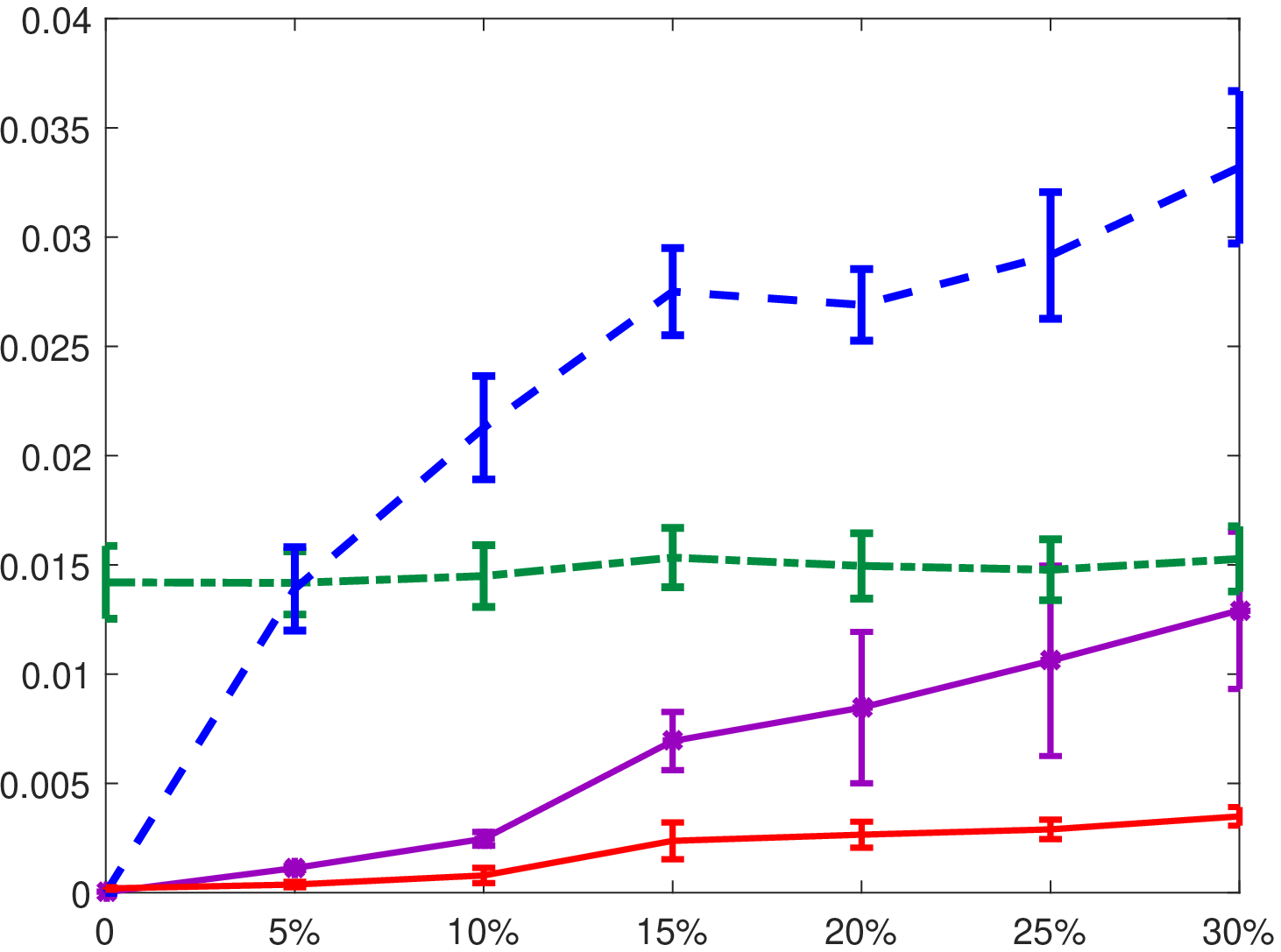}}
\subfigure[$\lambda=5$]{\includegraphics[width=0.45\textwidth]{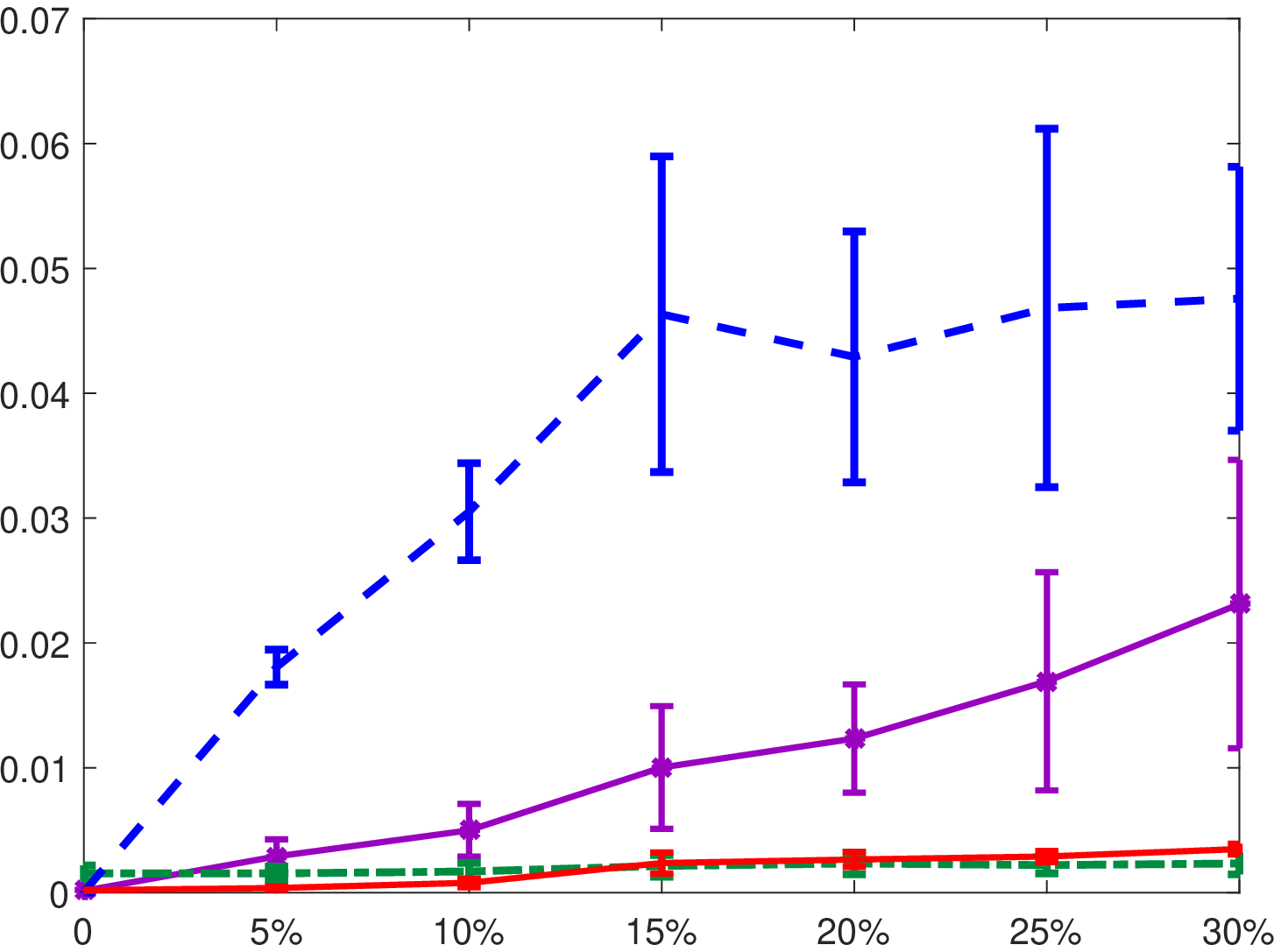}}
\caption{Test RMSE comparisons on the case study between four algorithms with different outlier percentage $\zeta$. $\lambda=0.1,0.5,1,5$ and $L=50$ are used in RVFL, Improved RVFL, and PRNNRW.}\label{fig:5}
\end{figure}
\begin{figure}[htbp!]
\centering
\subfigure[RVFL]{\includegraphics[width=0.42\textwidth]{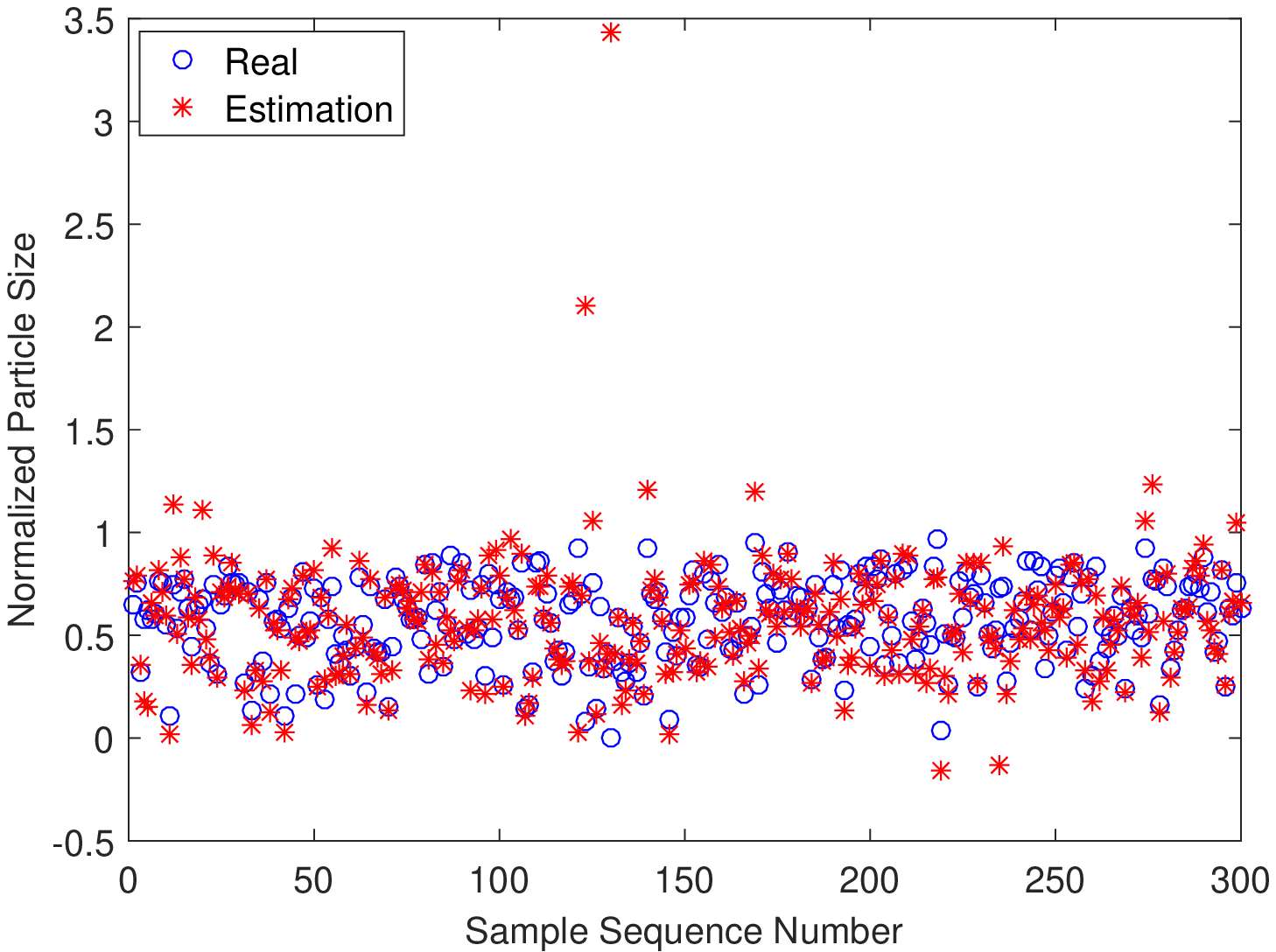}}
\subfigure[Improved RVFL]{\includegraphics[width=0.42\textwidth]{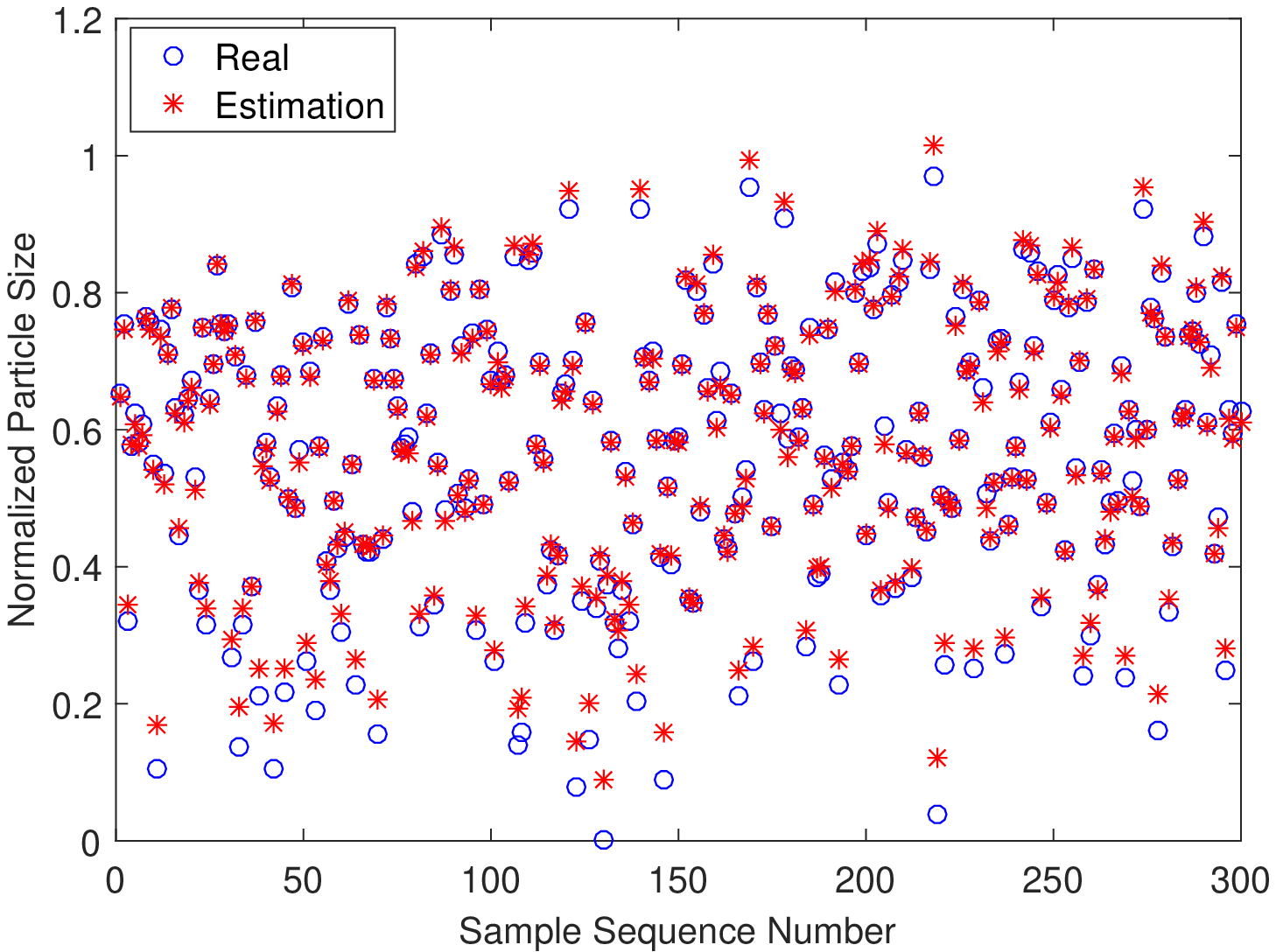}}
\subfigure[PRNNRW]{\includegraphics[width=0.42\textwidth]{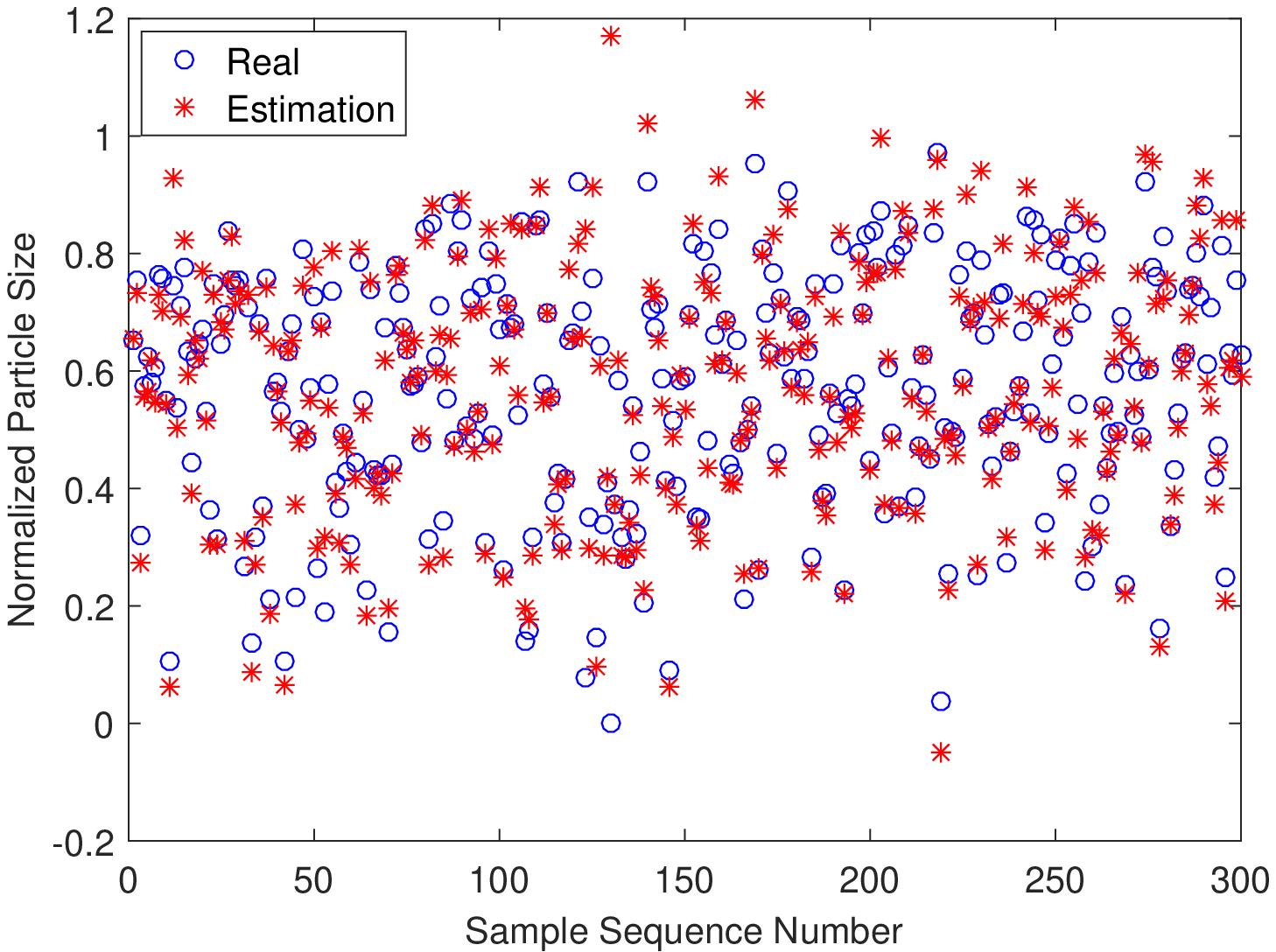}}
\subfigure[RSC-KDE]{\includegraphics[width=0.42\textwidth]{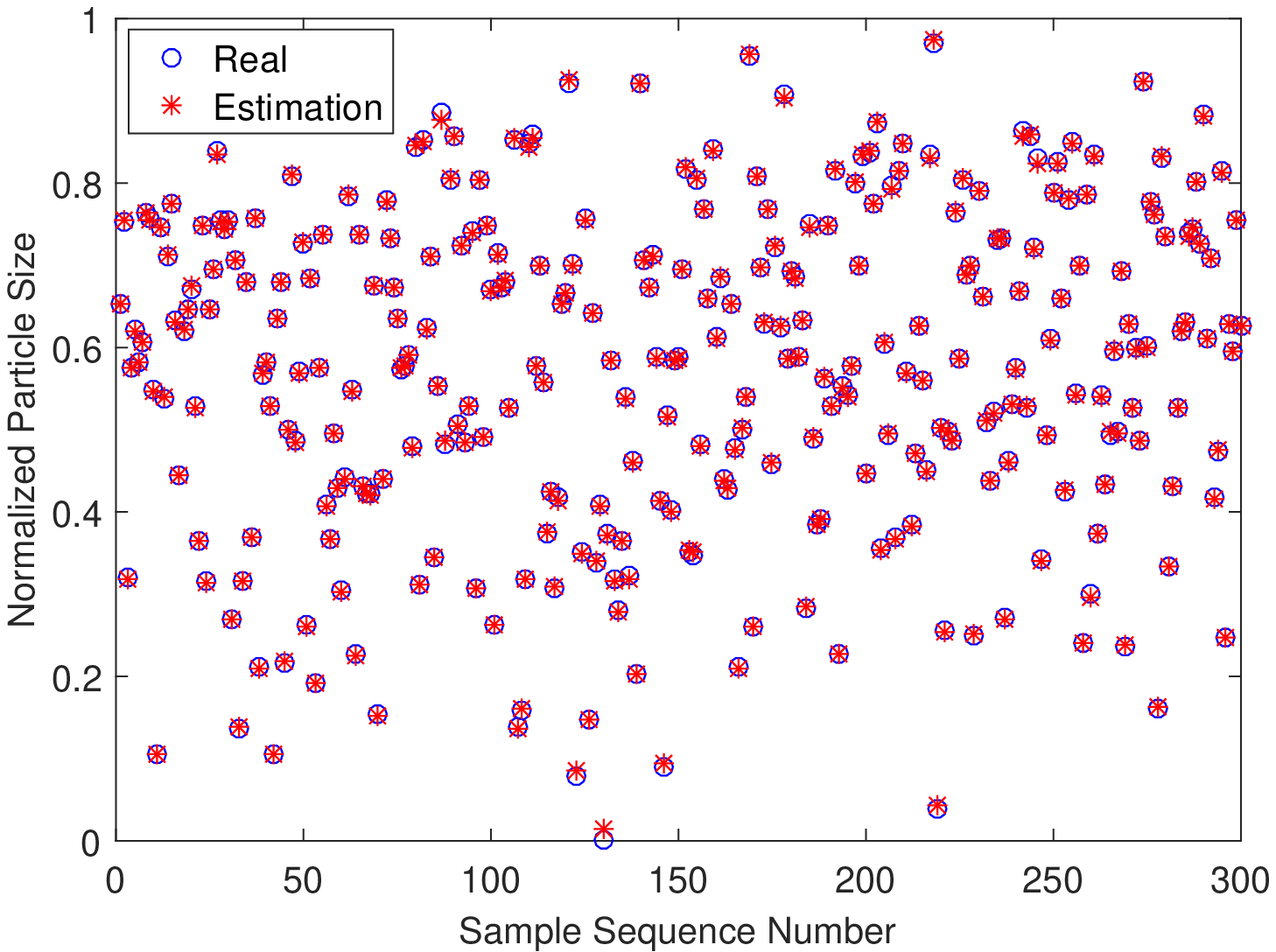}}
\caption{Test results of the four algorithms at $\zeta=30\%$ on the case study. $\lambda=1$ and $L=50$ are used in RVFL (a), Improved RVFL (b), and PRNNRW (c).}\label{fig:6}
\end{figure}

\begin{table}[!h]
\centering
\footnotesize
{\caption{Performance comparisons on the case study}\label{table4}}
\begin{center}
\begin{tabular}{ccccccc}\hline
\multirow{2}*{Scope Setting} &\multirow{2}*{Algorithm}&\multicolumn{4}{c}{Test Performance at Different Outlier Percentage (MEAN,STD)} \\
\cline{3-6}
 & &10$\%$&15$\%$&20$\%$&25$\%$ \\
\hline
\multirow{4}*{$\lambda=0.1$}&   RVFL  &0.0213,0.0027 &0.0279,0.0025 &0.0274,0.0025 &0.0288,0.0030 \\
&Improved RVFL&0.0018,0.0003 &0.0033,\textbf{0.0003} &0.0066,0.0006 &0.0097,0.0010 \\
&PRNNRW&0.1437,0.0063 &0.1451,0.0053 &0.1469,0.0053 &0.1465,0.0048 \\
&RSC-KDE&\textbf{0.0006},\textbf{0.0003} &\textbf{0.0020},0.0007 &\textbf{0.0026},\textbf{0.0006} &\textbf{0.0028},\textbf{0.0004}\\
\hline
\multirow{4}*{$\lambda=0.5$}&   RVFL     &0.0217,0.0026 &0.0278,0.0024 &0.0274,0.0022 &0.0295,0.0035\\
&Improved RVFL&0.0023,0.0003 &0.0064,0.0011 &0.0078,0.0010 &0.0103,0.0032 \\
&PRNNRW&0.0238,0.0005 &0.0238,\textbf{0.0005} &0.0234,0.0006 &0.0230,0.0005 \\
&RSC-KDE&\textbf{0.0006},\textbf{0.0003} &\textbf{0.0020},0.0007 &\textbf{0.0026},\textbf{0.0006} &\textbf{0.0028},\textbf{0.0004} \\
\hline
\multirow{4}*{$\lambda=1$}&   RVFL   &0.0213,0.0024 &0.0275,0.0020 &0.0269,0.0016 &0.0292,0.0029 \\
&Improved RVFL&0.0025,0.0003 &0.0069,0.0013 &0.0085,0.0035 &0.0106,0.0044 \\
&PRNNRW&0.0145,0.0014 &0.0153,0.0014 &0.0150,0.0015 &0.0148,0.0014 \\
&RSC-KDE&\textbf{0.0006},\textbf{0.0003} &\textbf{0.0020},\textbf{0.0007} &\textbf{0.0026},\textbf{0.0006} &\textbf{0.0028},\textbf{0.0004} \\
\hline
\multirow{4}*{$\lambda=5$}&   RVFL   &0.0305,0.0039 &0.0463,0.0126 &0.0429,0.0100 &0.0468,0.0144 \\
&Improved RVFL&0.0050,0.0021 &0.0100,0.0049 &0.0123,0.0043 &0.0169,0.0087 \\
&PRNNRW&0.0017,0.0007 &0.0021,0.0009 &\textbf{0.0023},0.0009 &\textbf{0.0022},0.0007 \\
&RSC-KDE&\textbf{0.0006},\textbf{0.0003} &\textbf{0.0020},\textbf{0.0007} &0.0026,\textbf{0.0006} &0.0028,\textbf{0.0004} \\
\hline
\end{tabular}
\end{center}
\end{table}

Before ending up this work, we conduct a robustness analysis on the key parameters ($L$ and $\nu$) to investigate their impacts on the performance of our proposed RSC-KDE algorithm. The test results for different combination of $L$ and $\nu$ are reported in Table 5 with $\zeta=0\%$, $\zeta=10\%$ and $\zeta=30\%$, respectively. For $\zeta=0\%$, there is no much difference among the results with different $\nu$ for each setting of $L$, implying that the AO process is not necessary. In this case, RSC-KDE is identical  to the original SC algorithm (SC-III in \cite{WangandLi_SCN}). When $\zeta=10\%$, the most appropriate setting of the architecture is $L=20$ while the iteration times in AO can be selected as $\nu=3,5,8,10,12$. At this percentage of outliers, it is fair to say that the accuracy of RSC-KDE with $L=20$ is preferable and stay within a stable level (i.e. RMSE is around 0.0010) provided that  $\nu$ is set equal or larger than 3. Similar to the case of $\zeta=30\%$, the most appropriate setting of the architecture is $L=10$ while the value of $\nu$ can be selected from the set $\{5,8,10,12\}$. All these records suggest that RSC-KDE performs robustly for uncertain data modelling with smaller iteration times in  AO (between 5 and 12). These empirical results  offer us some information on the setting of $\nu$, although it is data dependent. 

\begin{table}[htbp!]
\centering
\footnotesize
{\caption{Robustness analysis of $\nu$ and $L$ with different outlier percentage $\zeta$ on the case study}\label{table5}}
\begin{center}
\begin{tabular}{c|c|cccccc}\hline
\multirow{2}*{Outlier Percentage}&\multirow{2}*{Number of AO}&\multicolumn{6}{c}{Test Performance with Different Setting of $L$ (Mean RMSE) } \\
\cline{3-8}
&&$L=10$ &$L=20$ &$L=30$ &$L=50$ &$L=60$ &$L=80$ \\
\hline
\multirow{6}*{$\zeta=0\%$}&$\nu=2$ &0.0025 &0.0003 &0.0003 &0.0002&0.0002&0.0002\\
&$\nu=3$ &0.0025 &0.0003 &0.0003 &0.0002 &0.0002&0.0002\\
&$\nu=5$ &0.0025 &0.0003 &0.0003 &0.0002 &0.0002&0.0002\\
&$\nu=8$ &0.0026 &0.0003 &0.0003 &0.0002 &0.0002&0.0002\\
&$\nu=10$&0.0026 &0.0003 &0.0003 &0.0002 &0.0002&0.0002\\
&$\nu=12$&0.0028 &0.0003 &0.0003 &0.0002 &0.0002&0.0002\\
\hline
\multirow{6}*{$\zeta=10\%$}&$\nu=2$&0.0026 &0.0030&0.0072 &0.0216 &0.0334&0.0501\\
&$\nu=3 $&0.0024&0.0010&0.0024 &0.0100 &0.0200&0.0444\\
&$\nu=5 $&0.0024&0.0009&0.0012 &0.0076&0.0128&0.0290\\
&$\nu=8 $&0.0022&0.0010 &0.0012&0.0055&0.0091&0.0228\\
&$\nu=10$&0.0023&0.0010 &0.0012 &0.0060 &0.0094&0.0194\\
&$\nu=12$&0.0025&0.0009 &0.0012 &0.0057 &0.0088&0.0244\\
\hline
\multirow{6}*{$\zeta=30\%$}&$\nu=2$ &0.0101 &0.0173 &0.0544 &0.0950 &0.1315&0.2322\\
&$\nu=3$&0.0051 &0.0081 &0.0347 &0.0932 &0.1493&0.2057\\
&$\nu=5$&0.0040 &0.0056&0.0161&0.0966 &0.1574&0.2182\\
&$\nu=8$&0.0039 &0.0051 &0.0102 &0.0870 &0.1604&0.2326\\
&$\nu=10$&0.0038 &0.0053 &0.0095 &0.1017 &0.1688&0.2366\\
&$\nu=12$&0.0041 &0.0050&0.0099 &0.0959 &0.1621&0.2308\\
\hline
\end{tabular}
\end{center}
\end{table}
\section{Concluding Remarks}
Uncertain data modelling problems appear in many real-world applications, it is significant to develop advanced machine learning techniques to achieve better modelling performance.  This paper proposes a robust version of stochastic configuration networks for problem solving. Empirical results reported in this work clearly indicate that the proposed RSCNs, as one of the extensions of our recently developed SCN framework in \cite{WangandLi_SCN}, have great potential in dealing with robust data regression problems. From the implementation perspective, our design methodology needs an assumption on the availability of a clean validation dataset, which helps to prevent the learner from over-fitting during the course of incrementally constructing stochastic configuration networks. In practice, however, such a hypothesis is not always applicable. Thus, further research on this topic is necessary.  A plenty  of  explorations are being expected, such as the use of various cost functions for evaluating the output weights, development of online version of RSCNs, and distributed RSCNs for large-scale data modelling. \\

\textbf{Acknowledgment}

The authors would like to thank Dr Wei Dai from China University of Mining Technology for his sharing the data used in the case study. 

\bibliographystyle{elsarticle-num-names}

%\bibliographystyle{elsarticle-num}

\begin{thebibliography}{}
\small

\bibitem[{Cao et~al.(2015)Cao, Ye, and Wang}]{Cao2015}
\bibinfo{author}{F.~Cao}, \bibinfo{author}{H.~Ye}, \bibinfo{author}{D.~Wang},
  \bibinfo{title}{A probabilistic learning algorithm for robust modeling using
  neural networks with random weights}, \bibinfo{journal}{Information Sciences}
  \bibinfo{volume}{313} (\bibinfo{year}{2015}) \bibinfo{pages}{62--78}.


\bibitem[{Chen and Jain(1994)}]{Chen1994}
\bibinfo{author}{D.~S. Chen}, \bibinfo{author}{R.~C. Jain}, \bibinfo{title}{A
  robust backpropagation learning algorithm for function approximation},
  \bibinfo{journal}{IEEE Transactions on Neural Networks,}
  \bibinfo{volume}{5}~(\bibinfo{number}{3}) (\bibinfo{year}{1994})
  \bibinfo{pages}{467--479}.


\bibitem[{Chuang et~al.(2002)Chuang, Su, Tsong, and Hsiao}]{Chuang2002}
\bibinfo{author}{C.~C. Chuang}, \bibinfo{author}{S.~F. Su},
  \bibinfo{author}{J.~Tsong}, \bibinfo{author}{C.~C. Hsiao},
  \bibinfo{title}{Robust support vector regression networks for function
  approximation with outliers}, \bibinfo{journal}{ IEEE
  Transactions on Neural Networks,} \bibinfo{volume}{13}~(\bibinfo{number}{6})
  (\bibinfo{year}{2002}) \bibinfo{pages}{1322--1330}.


\bibitem[{Dai et~al.(2015)Dai, Liu, and Chai}]{Dai2015}
\bibinfo{author}{W.~Dai}, \bibinfo{author}{Q.~Liu}, \bibinfo{author}{T. Y.~Chai},
  \bibinfo{title}{Particle size estimate of grinding processes using random
  vector functional link networks with improved robustness},
  \bibinfo{journal}{Neurocomputing,} \bibinfo{volume}{169}
  (\bibinfo{year}{2015}) \bibinfo{pages}{361--372}.

\bibitem[{El-Melegy(2013)}]{El2013random}
\bibinfo{author}{M.~T. El-Melegy}, \bibinfo{title}{Random sampler M-estimator
  algorithm with sequential probability ratio test for robust function
  approximation via feed-forward neural networks}, \bibinfo{journal}{IEEE Transactions on Neural
  Networks and Learning Systems,}
  \bibinfo{volume}{24}~(\bibinfo{number}{7}) (\bibinfo{year}{2013})
  \bibinfo{pages}{1074--1085}.

\bibitem[{Gorban et al(2016)}]{Tyukin2009}
\bibinfo{author}{ A.~N. Gorban}, \bibinfo{author}{I. Tyukin}, \bibinfo{author}{D.~V. Prokhorov}, \bibinfo{author}{K. Sofeikov},
  \bibinfo{title}{Approximation with random bases: Pro et contra},  \bibinfo{journal}{Information Sciences} \bibinfo{volume}{364--365} (\bibinfo{year}{2016})  \bibinfo{pages}{129--145}.

\bibitem[{Hampel et~al.(2011)Hampel, Ronchetti, Rousseeuw, and
  Stahel}]{Hampel2011robust}
\bibinfo{author}{F.~R. Hampel}, \bibinfo{author}{E.~M. Ronchetti},
  \bibinfo{author}{P.~J. Rousseeuw}, \bibinfo{author}{W.~A. Stahel},
  \bibinfo{title}{Robust Statistics: The Approach Based on Influence
  Functions}, \bibinfo{publisher}{John Wiley \&
  Sons}, \bibinfo{year}{2011}.

\bibitem[{Huber and Ronchetti(1981)}]{Huber1981robust}
\bibinfo{author}{P.~Huber},
  \bibinfo{title}{Robust Statistics}, \bibinfo{journal}{Wiley Series in
  Probability and Mathematical Statistics}, \bibinfo{publisher}{Wiley},
  \bibinfo{year}{1981}.


\bibitem[{Igelnik and Pao(1995)}]{Igelnik1995}
\bibinfo{author}{B.~Igelnik}, \bibinfo{author}{Y.~H. Pao},
  \bibinfo{title}{Stochastic choice of basis functions in adaptive function
  approximation and the functional-link net}, \bibinfo{journal}{IEEE Transactions on Neural
  Networks,} \bibinfo{volume}{6}~(\bibinfo{number}{6})
  (\bibinfo{year}{1995}) \bibinfo{pages}{1320--1329}.

  \bibitem[{Latecki et~al.(2007)Latecki, Lazarevic, and Pokrajac}]{kde1}
\bibinfo{author}{L.~J. Latecki}, \bibinfo{author}{A.~Lazarevic},
  \bibinfo{author}{D.~Pokrajac}, \bibinfo{title}{Outlier detection with kernel
  density functions}, in: \bibinfo{booktitle}{Proceedings of the 5th International Conference on Machine Learning and Data Mining in Pattern Recognition}, \bibinfo{organizer}{Berlin, Heidelberg}, \bibinfo{year}{(2007)} \bibinfo{pages}{61--75}.


\bibitem[{Leroy and Rousseeuw(1987)}]{Leroy1987robust}
\bibinfo{author}{A.~M. Leroy}, \bibinfo{author}{P.~J. Rousseeuw},
  \bibinfo{title}{Robust Regression and Outlier Detection},
  \bibinfo{journal}{Wiley Series in Probability and Mathematical Statistics}, \bibinfo{publisher}{Wiley}, \bibinfo{year}{1987}.

\bibitem[{Liano(1996)}]{Liano1996}
\bibinfo{author}{K.~Liano}, \bibinfo{title}{Robust error measure for supervised
  neural network learning with outliers}, \bibinfo{journal}{IEEE Transactions on Neural Networks,}
   \bibinfo{volume}{7}~(\bibinfo{number}{1})
  (\bibinfo{year}{1996}) \bibinfo{pages}{246--250}.

\bibitem{LiandWang2016}
M. Li, D. Wang, Insights into randomized algorithms for neural networks:
practical issues and common pitfalls, Information Sciences 382-383 (2017) 170-178.

\bibitem[{Lowe(1988)}]{Broomhead1988}
\bibinfo{author}{D.~Lowe}, \bibinfo{title}{Multi-variable functional
  interpolation and adaptive networks}, \bibinfo{journal}{Complex Systems}
  \bibinfo{volume}{2} (\bibinfo{year}{1988}) \bibinfo{pages}{321--355}.



%\bibitem[{Mahoney(2011)}]{Mahoney2011}
%\bibinfo{author}{M.~W. Mahoney}, \bibinfo{title}{Randomized algorithms for
%  matrices and data}, \bibinfo{journal}{Foundations and Trends
%  in Machine Learning} \bibinfo{volume}{3}~(\bibinfo{number}{2})
%  (\bibinfo{year}{2011}) \bibinfo{pages}{123--224}.


\bibitem[{Meer et~al.(1991)Meer, Mintz, Rosenfeld, and Kim}]{Meer1991robust}
\bibinfo{author}{P.~Meer}, \bibinfo{author}{D.~Mintz},
  \bibinfo{author}{A.~Rosenfeld}, \bibinfo{author}{D.~Y. Kim},
  \bibinfo{title}{Robust regression methods for computer vision: A review},
  \bibinfo{journal}{International Journal of Computer Vision}
  \bibinfo{volume}{6}~(\bibinfo{number}{1}) (\bibinfo{year}{1991})
  \bibinfo{pages}{59--70}.


%
%\bibitem[{Pao et~al.(1994)Pao, Park, and Sobajic}]{Pao1994}
%\bibinfo{author}{Y. H. Pao}, \bibinfo{author}{G. H. Park},
%  \bibinfo{author}{D.~J. Sobajic}, \bibinfo{title}{Learning and generalization
%  characteristics of the random vector functional-link net},
%  \bibinfo{journal}{Neurocomputing} \bibinfo{volume}{6}~(\bibinfo{number}{2})
%  (\bibinfo{year}{1994}) \bibinfo{pages}{163--180}.
%
\bibitem[{Pao and Takefji(1992)}]{Pao1992}
\bibinfo{author}{Y. H. Pao}, \bibinfo{author}{Y.~Takefji},
  \bibinfo{title}{Functional-link net computing}, \bibinfo{journal}{IEEE
  Computer Journal} \bibinfo{volume}{25}~(\bibinfo{number}{5})
  (\bibinfo{year}{1992}) \bibinfo{pages}{76--79}.


\bibitem[{Ramsay and Scott(1993)}]{kde2}
\bibinfo{author}{P.~Ramsay}, \bibinfo{author}{D.~Scott},
  \bibinfo{title}{Multivariate Density Estimation, Theory, Practice, and
  Visualization}, \bibinfo{publisher}{Wiley}, \bibinfo{year}{1993}.


\bibitem[{Scardapane and Wang (2017)}]{ScardapaneandWang2017}
\bibinfo{author}{S.~Scardapane}, \bibinfo{author}{D.~Wang},
\bibinfo{title}{Randomness in neural networks: an overview}, \bibinfo{journal}{Wiley Interdisciplinary Reviews: Data Mining and Knowledge Discovery}
  \bibinfo{volume}{7}~(\bibinfo{number}{2}) (\bibinfo{year}{2017})
  \bibinfo{pages}{e1200. doi:10.1002/widm.1200}.

\bibitem[{Suykens et~al.(2002)Suykens, De~Brabanter, Lukas, and
  Vandewalle}]{Suykens2002}
\bibinfo{author}{J.~A. Suykens}, \bibinfo{author}{J.~De~Brabanter},
  \bibinfo{author}{L.~Lukas}, \bibinfo{author}{J.~Vandewalle},
  \bibinfo{title}{Weighted least squares support vector machines: robustness
  and sparse approximation}, \bibinfo{journal}{Neurocomputing}
  \bibinfo{volume}{48}~(\bibinfo{number}{1}) (\bibinfo{year}{2002})
  \bibinfo{pages}{85--105}.

\bibitem[{Torr and Zisserman(2000)}]{Torr2000mlesac}
\bibinfo{author}{P.~H. Torr}, \bibinfo{author}{A.~Zisserman},
  \bibinfo{title}{MLESAC: A new robust estimator with application to estimating
  image geometry}, \bibinfo{journal}{Computer Vision and Image Understanding}
  \bibinfo{volume}{78}~(\bibinfo{number}{1}) (\bibinfo{year}{2000})
  \bibinfo{pages}{138--156}.


\bibitem[{Wang and Li(2017)}]{WangandLi_SCN}
\bibinfo{author}{D.~Wang}, \bibinfo{author}{M.~Li}, \bibinfo{title}{Stochastic
  configuration networks: Fundamentals and algorithms},
  \bibinfo{journal}{arXiv:1702.03180 [cs.NE]}, 10 Feb, 2017.

\bibitem[{Zhuang et~al.(1992)Zhuang, Wang, and Zhang}]{Zhuang1992highly}
\bibinfo{author}{X.~Zhuang}, \bibinfo{author}{T.~Wang},
  \bibinfo{author}{P.~Zhang}, \bibinfo{title}{A highly robust estimator through
  partially likelihood function modeling and its application in computer
  vision}, \bibinfo{journal}{IEEE Transactions on Pattern Analysis and Machine
  Intelligence} \bibinfo{volume}{14}~(\bibinfo{number}{1}) (\bibinfo{year}{1992})
  \bibinfo{pages}{19--35}.

\end{thebibliography}

\end{document}